\documentclass{ieeeaccess}
\usepackage{cite}
\usepackage{amsmath,amssymb,amsfonts}
\usepackage{algorithmic}
\usepackage{algorithm}
\usepackage{graphicx}
\usepackage{caption}
\usepackage{url}
\usepackage{multirow}
\usepackage{textcomp}
\usepackage{float,lscape}
\usepackage{multirow}
\usepackage{multicol}
\usepackage{booktabs}
\def\BibTeX{{\rm B\kern-.05em{\sc i\kern-.025em b}\kern-.08em
		T\kern-.1667em\lower.7ex\hbox{E}\kern-.125emX}}
\begin{document}
	\history{Date of publication xxxx 00, 0000, date of current version xxxx 00, 0000.}
	\doi{10.1109/ACCESS.2017.DOI}
	\title{Optimal trees selection for classification via out-of-bag assessment and sub-bagging}
	\author{\uppercase{Zardad Khan}\authorrefmark{1,2},\uppercase{Naz Gul}\authorrefmark{1}, \uppercase{Nosheen Faiz}\authorrefmark{1,2}, \uppercase{Asma Gul}\authorrefmark{3}, \uppercase{Werner Adler}\authorrefmark{4}, \uppercase{Berthold Lausen}\authorrefmark{2,4} }
	\address[1]{Department of Statistics, Abdul Wali Khan University Mardan, 23200 Pakistan}
	\address[2]{Department of Mathematical Sciences, University of Essex, UK}
		\address[3]{Department of Statistics, Shaheed Benazir Bhutto Women University Peshawar, Pakistan}
		\address[4]{Department of Biometry and Epidemiology, University of Erlangen-Nuremberg, Germany}	
	\corresp{Corresponding authors: Berthold Lausen (e-mail: blausen@essex.ac.uk) \& Zardad Khan ( e-mail:zardadkhan@awkum.edu.pk)}

\tfootnote{``We acknowledge support from grant number ES/L011859/1, from The Business and Local Government Data Research Centre, funded by the Economic and Social Research Council to provide researchers and analysts with secure data services.''}


\begin{abstract}
	The effect of training data size on machine learning methods has been well investigated over the past two decades. The predictive performance of tree based machine learning methods, in general, improves with a decreasing rate as the size of training data increases. We investigate this in optimal trees ensemble (OTE) where the method fails to learn from some of the training observations due to internal validation. Modified tree selection methods are thus proposed for OTE to cater for the loss of training observations in internal validation. In the first method, corresponding out-of-bag (OOB) observations are used in both individual and collective performance assessment for each tree. Trees are ranked based on their individual performance on the OOB observations. A certain number of top ranked trees is selected and starting from the most accurate tree, subsequent trees are added one by one and their impact is recorded by using the OOB observations left out from the bootstrap sample taken for the tree being added. A tree is selected if it improves predictive accuracy of the ensemble. In the second approach, trees are grown on random subsets, taken without replacement-known as sub-bagging, of the training data instead of bootstrap samples (taken with replacement). The remaining observations from each sample are used in both individual and collective assessments for each corresponding tree similar to the first method. Analysis on 21 benchmark datasets and simulations studies show improved performance of the modified methods in comparison to OTE and other state-of-the-art methods.
\end{abstract}

\titlepgskip=-15pt

\maketitle
\section{Introduction}
Ensemble techniques help to improve machine learning results by integrating multiple models. Using ensemble methods allows to produce better predictions compared to a single base model. There is a huge literature on ensemble methods which is fast growing \cite{quintian2020novel,yang2020co,wang2020three,ali2020k,jia2020research}. One of the most widely used ensemble method is random forest \cite{breiman2001random} that combines classification and regression trees \cite{breiman1984classification,Breiman1996} as the base model.
Classification and regression tree, the building block of many tree based ensemble methods, including random forest, depends both on the quality and quantity of training data \cite{sebbanu2000impact}. A tree grown with more meaningful information (data points) will give better results than the one built otherwise \cite{sebbanu2000impact}.

The efficacy of combining a large number of individual classifiers, also called base learners, has been well studied \cite{schapire1990strength,domingos1996using,quinlan1996bagging,maclin2011popular,hothorn2003double,gul2016ensemble,lausser2016rank}. The main advantage of combining the results of many variants of the same classifier is that it leads to a reduction in the generalization error of the resultant ensemble classifier \cite{domingos1996using,quinlan1996bagging,bauer1999empirical,maclin2011popular,tzirakis2017t3c}. The reason behind this is that the variants of the same classifier have different inductive biases. This kind of diversity results in a reduction of variance-error without increasing the bias-error \cite{mitchell1997machine,tumer1996error,ali1996error}. Following this, Breiman \cite{breiman2001random} argued that diverse and individually strong classifiers will result in an efficient ensemble, while proposing his famous random forest method. Breiman achieved this by selecting $p < d$ features at each node while growing trees on bootstrap samples. The random forest algorithm has been extensively used in solving various classification and regression problems related to medicine \cite{zhu2018class}, banking and finance \cite{lin2017ensemble}, engineering \cite{kim2019fast}, etc. and has attracted a significant attention of the research community.  For further diversity and improvement in tree ensembles, Khan et al. \cite{khan2016ensemble,khan2020ensemble} proposed selecting the most accurate trees based on their performance on out-of-bag (OOB) observation. These trees were then further assessed for their collective performance using a subset of the training data as internal validation data for final ensemble. They called this method optimal trees ensemble (OTE). OTE not only showed improved predictive accuracy in comparison to several other state-of-the-art methods, but also reduced ensemble size. 

However, while selecting the optimal trees, trees in OTE fail to learn from some of the training observations due to the internal validation. This paper suggests modified methods of tree selection to avoid this issue. In the first method, corresponding out-of-bag (OOB) observations are used in both individual and collective performance assessment for each tree. Trees are ranked based on their individual performance on the OOB observations. A certain number of top ranked trees is selected and starting from the most accurate tree, subsequent trees are added one by one and their impact is recorded by using the OOB observations left out from the bootstrap sample taken for the tree being added. A tree is selected if it improves predictive accuracy of the ensemble. In the second approach, trees are grown on random subsets, taken without replacement,  of the training data instead of bootstrap samples. The remaining observations from each sample are used in both individual and collective assessments for each corresponding tree similar to the first method. Using 21 benchmark problems, the results from the new approaches are compared with those of $k$NN, tree classifier, random forest, node harvest, support vector machine, random projection ensemble and OTE. The methods are further assessed by using the simulation models given in \cite{khan2016ensemble} by generating datasets of two different sizes. The remainder of the paper is arranged as follows. The proposed modified approaches, their algorithms and some other related methods are given in Section \ref{method}, experiments and findings based on simulated and benchmark data sets are given in Section \ref{exp}. Conclusion based on the work done in the article is given in Section \ref{conclusion}.
\section{Optimal Tree Selection} \label{method}
Using the notation of \cite{khan2016ensemble}, let $\mathcal{L}=({\mathbf{X}},{{Y}})=\{({\mathbf{x}}_1,y_1),({\mathbf{x}}_2,y_2),...,({\mathbf{x}}_n,y_n)\}$ be the given training data, where $\mathbf{X}$ is an $n \times d$ matrix and $Y$ a vector of length $n$. . The ${\mathbf{x}_i}$ are instances on $d$ features and $y_i$ are binary values representing two possible classes. OTE partitions $\mathcal{L}=({\mathbf{X}},{{Y}})$ randomly into two parts, $\mathcal{L_B}=({\mathbf{X}_B},{{Y_B}})$ and $\mathcal{L_O}=({\mathbf{X}_O},{{Y_O}})$. The steps of OTE are given as
\begin{enumerate}
	\item Trees are developed on $T$ bootstrap samples from $\mathcal{L_B}=({\mathbf{X}_B},{{Y_B}})$, using the random forest approach.
	\item The grown trees are ranked in ascending order of their prediction error on out-of-bag data and $M$ top ranked trees are taken.
	\item Starting from the highest ranked tree, the $M$ selected trees are added one by one and $\mathcal{L_O}=({\mathbf{X}_O},{{Y_O}})$ is applied to see whether the added tree improves predictive accuracy. A tree is selected if it improves accuracy and is discarded otherwise.
	\item The selected trees are integrated together for the final ensemble that is used for predicting new/test data.
\end{enumerate}
Although OTE has achieved improved performance as compared to the other methods on the given benchmark and simulated datasets as shown in \cite{khan2016ensemble},
a problem arises when there is a small number of observations in the data. As the trees are grown on a subset of the training data leaving the remaining observations, say V\%, for internal validation, this might result in missing out some useful information to learn from during the process of growing the trees and increases the variance of the classifier \cite{efron1997improvements,adler2009bootstrap}. It has been investigated that classification tree strongly depends on the amount of information present in the training data \cite{sebbanu2000impact}. To utilise the whole training data while growing and selecting optimal trees, two approaches are proposed in this paper. 
\subsection{Out-of-bag assessment}
In this method out-of-bag (OOB) observations are used in both individual and ensemble assessment of the trees. In bootstrapping, as the samples are taken with replacement, some observation are repeated and some are left out from the samples. Studies show that while bootstrapping, about 1/3 of the total training data are left out from the samples \cite{efron1993introduction}. These are called out-of-bag (OOB) observations and play no role in growing classification trees. They can rather be used in assessing the predictive ability of the trees and statistic values thus produced are called OOB estimates. Let $S_t$, $t=1,\ldots,T$ be the bootstrap sample and $\bar{S_t}$ be the corresponding OOB sample; $H(S_t)$ is the classification tree grown on $S_t$. Also suppose that $\widehat{err}_t$ is the error of $H(S_t)$ on $\bar{S_t}$ called the OOB error. About  37\% of the observations in the training  set  $\mathcal{L}$ do not appear in a particular bootstrap sample $S_t$. These observations can thus be used as unseen test examples. The steps of the proposed method under this approach are:
\begin{enumerate}
	\item Grow $T$ classification trees by the method of random forest on $S_t, t=1,\ldots,T$. Estimate $\widehat{err}_t$ for each tree as
	\begin{equation}
	\widehat{err}_t= \frac{1}{|\bar{S_t}|}\sum_{\mathbf{x_i}\in \bar{S_t}}I(y\ne \hat{y}),
	\end{equation}
	where $y$ is the true class label in the bootstrap sample $\bar{S_t}$, $\hat{y}$ is the corresponding estimated value by tree $H(S_t)$ and $|\bar{S_t}|$ is the size of the OOB sample. $I(y\ne \hat{y})$ is an indicator function with values 0 or 1 given as
	\begin{equation}
	I(y\ne \hat{y})=
	\begin{cases}
	1, & \text{if}\ y \ne \hat{y}, \\
	0, & \text{otherwise.}
	\end{cases}
	\end{equation}
	\item Arrange the trees for ranking in ascending order with respect to $\widehat{err}_t$; select the top ranked $M$ trees. Let $H^{R_1}(.),\ldots H^{R_M}(.)$, be the highest, second highest and so on, ranked trees.   
	\item Starting from $H^{R_1}(.)$,  test consecutive $H^{R_j}(.), j=2,\ldots, M$ one by one by using the corresponding OOB observations as the test data. Select $H^{R_j}(.)$ if
	\begin{equation}
		{\hat{\mathcal{BS}}}^{\langle j+ \rangle} < {\hat{\mathcal{BS}}}^{\langle j- \rangle},
	\end{equation}
	where  ${\hat{\mathcal{BS}}}^{\langle j- \rangle}$ is the Brier score \cite{brier1950verification} calculated for the ensemble not having the $j$th tree and ${\hat{\mathcal{BS}}}^{\langle j+ \rangle}$ is the Brier score of the method including the $j$th tree. An estimator for the Brier score is given as
	\begin{equation}
		\hat{\mathcal{BS}} =  \frac{\sum_{i=1}^{\text{\# of test observations}}\left(y_i-\hat{P}(y_i | {\bf{X}})\right)^2}{\text{total \# of test observations}},
	\end{equation}
	$y_i$ is the state of the class value for observation $i$ in the $(0,1)$ form and $\hat{P}(y | {\bf{X}})$ is the response/class probability estimate of the method given the variables.
	\item Integrate the trees for predicting new/test data.
\end{enumerate}		
\subsection{Sub-sampling/sub-bagging based assessment}
Under this approach, random sub-samples without replacement from the training data $\mathcal{L}=({\bf{X}},{\bf{Y}})$ are taken for growing the trees. The remaining observations from each sample are used as the test data for assessing the predictive performance of each corresponding tree, in contrary to using the OOB observations.  Let $\mathcal{S}_t$, $t=1,\ldots,T$ be the random sample of size $m < n$, $n$ being the number of instances in the training data, and $\bar{\mathcal{S}_t}$ be the corresponding remaining subset of observations of size $n-m$; $H(\mathcal{S}_t)$ is the classification tree grown on $\mathcal{S}_t$. Also suppose that $\widehat{err\_sub}_t$ is the error of $H(\mathcal{S}_t)$ on $\bar{\mathcal{S}_t}$. Then the steps of the proposed method under this approach are:
\begin{enumerate}
	\item Grow $T$ classification trees on $\mathcal{S}_t, t=1,\ldots,T$. Estimate $\widehat{err\_sub}_t$ for each tree using $\bar{\mathcal{S}_t}$.
	\item Rank the trees in ascending order with respect to $\widehat{err\_sub}_t$; select the top ranked $M$ trees. Let $H^{R_1}(.),\ldots H^{R_M}(.)$, be the highest, second highest and so on, ranked trees.   
	\item Starting from $H^{R_1}(.)$,  test consecutive $H^{R_j}(.), j=2,\ldots, M$ one by one by using the corresponding observations in the sample remainder as the test data. Select $H^{R_j}(.)$ based on the criteria used in Step 3 of the method in previous section.
	\item Integrate the trees for predicting new/test data.
\end{enumerate}
Both of the above methods are inspired from Breiman's \cite{breiman2001random} upper bound defined for the overall prediction error $PE^*$ of random forest algorithm given as
\begin{equation}\label{bound}
PE^* \leq \bar{\rho} \, PE_t.
\end{equation}
In Equation, \ref{bound} $t = 1,2,3,...,T$ where $T$ is the total number of trees grown in the forest, $\bar{\rho}$ is the weighted correlation between residuals from two independent classification trees calculated as the mean (expected) value of their correlation over entire random forest and $PE_t$ is the estimated prediction error of some $t$th tree in the forest.	

\begin{figure}
	\centering
	\begingroup\setlength{\fboxsep}{-26pt}%
	\fbox{
	\includegraphics[width=10cm]{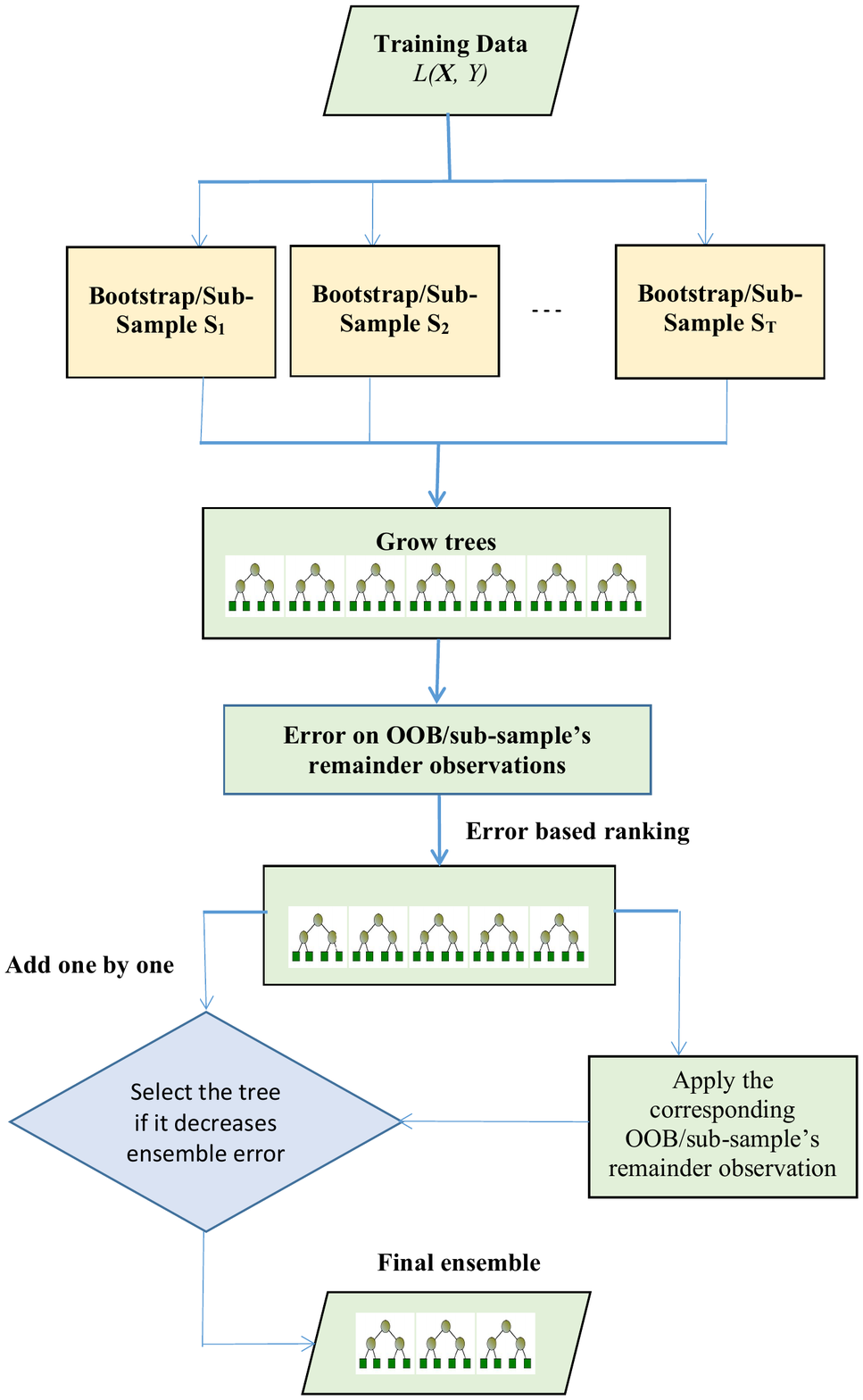}}\endgroup
	\caption{Flow chart of the proposed ensembles}
	\label{flow_chart}
	\end{figure}

A flowchart showing the general work flow of the proposed ensembles in given in Figure \ref{flow_chart}.
Care should be taken for deciding on the size $m$ of sample drawn for growing trees under this approach in relation to the total number of observations in $\mathcal{L}$. This is necessary for avoiding potentially redundant trees in the forest in that there can be only $n\choose m$ combinations of the training data to grow trees. As the final ensemble selects only a small number of diverse and accurate trees, this approach might be very helpful in small data situations where only a few trees are needed and missing more observations from sample is costly. This approach is expected to work similar to the OOB assessment method when $n-m$ is chosen to be $2/3$ of the training data. Similar study illustrating this has been done in \cite{sebbanu2000impact}.  

These approaches are novel in the following sense:
\begin{itemize}
	\item The proposed methods investigate optimal tree selection without loosing informative training data.
	\item The method based on sub-bagging tries to allocate more training data as compared to the out-of-bag assessment. This approach keeps 10\% of the given training data for internal validation and the remaining 90\% of the data is used for growing the trees. 
	\item The tree selection approaches proposed in the paper are based on individual accuracy of the base tree classifiers as well as their diversity in the ensemble in addition to minimizing the loss of informative information in the learning process. 
	\item Based on the above intuitions, the proposed methods could effectively be used in small data situations for optimal trees selection. 
\end{itemize}  

\subsection{Other related work}
Several methods are available in literature that are based on the idea of tree selection from bagged tree forest. These methods are based on bagging or its variant in that they improve on unstable estimators or classifiers. Methods based on bagging are useful especially for high dimensional data set problems. B\"uhlmann and Bin \cite{buhlmann2002analyzing} formalized the idea of instability and derived theoretical results to analyze the variance minimization effect of bagging (or the variants). To do this, B\"uhlmann and Bin \cite{buhlmann2002analyzing} considered hard decision problems including estimation after testing in regression and decision trees for  classifiers regression functions. They argued that hard decisions create instability, and bagging is helpful in smoothing such hard decisions which results in smaller variance and mean squared error \cite{buja2000effect}. B\"uhlmann and Bin \cite{buhlmann2002analyzing} motivated sub-bagging based on sub-sampling as an alternative the conventional aggregation scheme by deriving theoretical explanation. Sub-bagging is shown as computationally cheaper with approximately the same accuracy as bagging. Bagging has led to a large pool of methods including random forest and other ensemble classifiers. Authors have further worked on reducing the size of bagging based ensemble methods.  
 Latinne et al. \cite{latinne2001limiting} proposed a method to avoid overproducing trees in the ensemble by determining the least number of classification trees that could give comparable results to a standard size ensemble. McNemar test of significance is used decide between forests with different number of trees based on their prediction error. Bernard et al. \cite{bernard2009selection} proposed the methods of sequential backward elimination and sequential forward selection methods to find sub-optimal forests. Li et al. \cite{li2010trees} proposed the idea of weighting the trees for random forest ensemble to learn data with large dimensions. They exploited out-of-bag observation for tree weighting in the forest. Adler et al. \cite{adler2016ensemble} have proposed ensemble pruning for solving class imbalanced problem by using Brier score and AUC for Glaucoma detection. Different number of trees for random forest were checked by Oshiro et al. \cite{oshiro2012many} so as to see after what point adding further trees results in no gain. They used 29 benchmark datasets to argue that after a certain number, adding further trees does not contribute to ensemble performance. Zhang and Wang \cite{zhang2009search} proposed the similarity based approach between the trees of the forest and suggested to remove trees that were similar. Khan et al. \cite{khan2016ensemble} proposed the idea of building an ensemble of probability estimation trees that are accurate and diverse and proposed to discard trees that are individually weak and do not contribute to ensemble. 
Based on a similar idea, ensembles selection for $k$NN classifiers has  been given where in addition to individual strength of classifiers, $k$NN models are build on different random subsets of the whole features set instead of using the entire features \cite{gul2016ensemble,gul2016ensemble2}.

\section{Experiments and results}\label{exp}
\subsection{Simulation} \label{simulation}
This section gives our analysis on simulated datasets using the simulation models porposed in \cite{khan2016ensemble}. The main idea of using these simulation models is to present slightly difficult recognition problems for simple classifiers like CART and $k$NN, and also to give a much challenging task for the most sophisticated classifiers like random forest and SVM. To this end, in all the four models, various complexity levels are taken by varying the weights $\lambda_{ijk}$ of the tree nodes. This gave four different values of the Bayes error for the models where the smallest error means that the dataset has meaningful structure and the highest Bayes error show that there are less/no meaningful structures. Various values of $\lambda_{ijk}$ used in Scenarios 1, 2, 3, and 4 are given in Table \ref{lambdas}. The corresponding node weights for each of the models to get various complexity levels are given in the columns of the table for $k=1,2,3,4$. Equation used for generating class membership probabilities of the binary class variable $\bf{Y} = $\text{ Bernoulli}$(p)$ given the $n \times 3T$ dimensional vector ${\bf{X}}$ of $n$ $iid$ observations from Uniform$(0,1)$ is
\begin{equation} \label{model}
	p(y|{\bf{X}}) = \frac{exp\left(\theta_2\times\left(\frac{\mathcal{P}_{m}}{T} - \theta_1\right)\right)}{1 + exp\left(\theta_2\times\left(\frac{\mathcal{P}_{m}}{T} - \theta_1\right)\right)}, \mbox{  where  } \mathcal{P}_{m} = \sum_{t=1}^T\hat{p}_{t}.
\end{equation}
$\theta_1$ and $\theta_2$ are arbitrary values, $m=1,2,3,4$ represents a scenario and $\mathcal{P}_{m}$'s are $n \times 1$ probability vectors. $T$ shows total number of trees in a scenario and $\hat{p}_{t}$'s are class probabilities for a binary response in $\bf{Y}$. The following structure generate the $\hat{p}_{t}$'s
\begin{eqnarray*}
	\hat{p}_{1} &=& \lambda_{11k} \times {\bf{1}}_{(x_1 \leq 0.5 \& x_3 \leq 0.5)} + \lambda_{12k} \times {\bf{1}}_{(x_1 \leq 0.5 \& x_3 > 0.5)}\\ &&+ \lambda_{13k} \times \mathbf{I}_{(x_1 > 0.5 \& x_2 \leq 0.5)}	 + \lambda_{14k} \times \mathbf{I}_{(x_1 > 0.5 \& x_2 > 0.5)},\\
	\hat{p}_{2} &=& \lambda_{21k} \times \mathbf{I}_{(x_4 \leq 0.5 \& x_6 \leq 0.5)} + \lambda_{22k} \times \mathbf{I}_{(x_4 \leq 0.5 \& x_6 > 0.5)} \\&&+ \lambda_{23k} \times \mathbf{I}_{(x_4 > 0.5 \& x_5 \leq 0.5)}
	+ \lambda_{24k} \times \mathbf{I}_{(x_4 > 0.5 \& x_5 > 0.5)},\\
	\hat{p}_{3} &=& \lambda_{31k} \times \mathbf{I}_{(x_7 \leq 0.5 \& x_8 \leq 0.5)} + \lambda_{32k} \times \mathbf{I}_{(x_7 \leq 0.5 \& x_8 > 0.5)}\\&& + \lambda_{33k} \times \mathbf{I}_{(x_7 > 0.5 \& x_9 \leq 0.5)}
	+ \lambda_{34k} \times \mathbf{I}_{(x_7 > 0.5 \& x_9 > 0.5)},\\
	\hat{p}_{4}&=& \lambda_{41k} \times \mathbf{I}_{(x_{10} \leq 0.5 \& x_{11} \leq 0.5)} + \lambda_{42k} \times \mathbf{I}_{(x_{10} \leq 0.5 \& x_{11} > 0.5)} \\&&+ \lambda_{43k} \times \mathbf{I}_{(x_{10} > 0.5 \& x_{12} \leq 0.5)}
	+ \lambda_{44k} \times \mathbf{I}_{(x_{10} > 0.5 \& x_{12} > 0.5)},\\
	\hat{p}_{5}&=& \lambda_{51k} \times \mathbf{I}_{(x_{13} \leq 0.5 \& x_{14} \leq 0.5)} + \lambda_{52k} \times \mathbf{I}_{(x_{13} \leq 0.5 \& x_{14} > 0.5)}\\&& + \lambda_{53k} \times \mathbf{I}_{(x_{13} > 0.5 \& x_{15} \leq 0.5)}
	+ \lambda_{54k} \times \mathbf{I}_{(x13 > 0.5 \& x_{15} > 0.5)},\\
	\hat{p}_{6} &=& \lambda_{61k} \times \mathbf{I}_{(x_{16} \leq 0.5 \& x_{17} \leq 0.5)} + \lambda_{62k} \times \mathbf{I}_{(x_{16} \leq 0.5 \& x_{17} > 0.5)}\\&& + \lambda_{63k} \times \mathbf{I}_{(x_{16} > 0.5 \& x_{18} \leq 0.5)} 
	+ \lambda_{64k} \times \mathbf{I}_{(x16 > 0.5 \& x_{18} > 0.5)},
\end{eqnarray*}
where $0 < \lambda_{ijk} < 1 $ are node weights in the trees, $k=1,2,3,4$ and $\mathbf{I}_{(condition)}$ is an indicator function that yields a $1$ if the stated condition is satisfied and $0$ if not . Note that the basic principle of random forest is followed while growing the trees by taking $p<d$ variables during nodes splitting. The various simulation scenarios are outlined as given below. 
\subsubsection{Scenario 1}
This is a relatively simple scenario consisting of $T=3$ tree components each with 3 variables, $\mathcal{P}_{1} = \sum_{t=1}^3\hat{p}_{t}$ and  ${\bf{X}}$ is a $n \times 9$ vector.
\subsubsection{Scenario 2}
This scenario has four tree components i.e. $T=4$ trees where $\mathcal{P}_{2} = \sum_{t=1}^4\hat{p}_{t}$ which follows that  ${\bf{X}}$ becomes a $n \times 12$ vector.
\subsubsection{Scenario 3}
This scenario has $T=5$ trees such that $\mathcal{P}_{3} = \sum_{t=1}^5\hat{p}_{t}$ and  ${\bf{X}}$ is a $n \times 15$ dimensional vector.
\subsubsection{Scenario 4}
This the most complex scenario having $T=6$ tree components following that, $\mathcal{P}_{4} = \sum_{t=1}^6\hat{p}_{t}$ and  ${\bf{X}}$ is a $n \times 18$ dimensional vector.


\onecolumn
\renewcommand{\arraystretch}{2}
\begin{table}[h!]
	\centering
	\caption{Node weights, $\lambda_{ijk}$, used in simulation scenarios. Tree number is shown by $i$, node number in each tree by $j$ and $k$ shows a variant of the weights to get the complexity levels in each scenario \cite{khan2016ensemble}.}
	\setlength{\tabcolsep}{3.5pt}
	\fontsize{9}{9}\selectfont
	\begin{tabular}{ccccccc|ccccccc|ccccccc|cccccc}
		\hline
		& \multicolumn{5}{c}{Scenario 1}           & \multicolumn{1}{c}{} & \multicolumn{6}{c}{Scenario 2}                   & \multicolumn{1}{c}{} & \multicolumn{6}{c}{Scenario 3}                   & \multicolumn{1}{c}{} & \multicolumn{6}{c}{Scenario 4} \\
		\midrule
		&       & \multicolumn{4}{c}{k}         & \multicolumn{1}{c}{} &       &       & \multicolumn{4}{c}{$k$}         & \multicolumn{1}{c}{} &       &       & \multicolumn{4}{c}{k}         & \multicolumn{1}{c}{} &       &       & \multicolumn{4}{c}{k} \\
		$i$     & \multicolumn{1}{c}{$j$} & 1     & 2     & 3     & 4     &       & \multicolumn{1}{c}{i} & \multicolumn{1}{c}{j} & 1     & 2     & 3     & 4     &       & \multicolumn{1}{c}{i} & \multicolumn{1}{c}{j} & 1     & 2     & 3     & 4     &       & \multicolumn{1}{c}{i} & \multicolumn{1}{c}{j} & 1     & 2     & 3     & 4 \\ \ \\
		\multicolumn{1}{c}{\multirow{4}[1]{*}{1}} & 1     & 0.9   & 0.8   & 0.7   & 0.6   &       & \multicolumn{1}{c}{\multirow{4}[1]{*}{1}} & 1     & 0.9   & 0.8   & 0.7   & 0.6   &       & \multicolumn{1}{c}{\multirow{4}[1]{*}{1}} & 1     & 0.9   & 0.9   & 0.9   & 0.8   &       & \multicolumn{1}{c}{\multirow{4}[1]{*}{1}} & 1     & 0.9   & 0.9   & 0.9   & 0.8 \\
		\multicolumn{1}{c}{} & 2     & 0.1   & 0.2   & 0.3   & 0.4   &       & \multicolumn{1}{c}{} & 2     & 0.1   & 0.2   & 0.3   & 0.4   &       & \multicolumn{1}{c}{} & 2     & 0.1   & 0.1   & 0.1   & 0.2   &       & \multicolumn{1}{c}{} & 2     & 0.1   & 0.1   & 0.1   & 0.2 \\
		\multicolumn{1}{c}{} & 3     & 0.1   & 0.2   & 0.3   & 0.4   &       & \multicolumn{1}{c}{} & 3     & 0.1   & 0.2   & 0.3   & 0.4   &       & \multicolumn{1}{c}{} & 3     & 0.1   & 0.1   & 0.1   & 0.2   &       & \multicolumn{1}{c}{} & 3     & 0.1   & 0.1   & 0.1   & 0.2 \\
		\multicolumn{1}{c}{} & 4     & 0.9   & 0.8   & 0.7   & 0.6   &       & \multicolumn{1}{c}{} & 4     & 0.9   & 0.8   & 0.7   & 0.6   &       & \multicolumn{1}{c}{} & 4     & 0.9   & 0.9   & 0.9   & 0.8   &       & \multicolumn{1}{c}{} & 4     & 0.9   & 0.9   & 0.9   & 0.8 \\
		\hline 
		\multicolumn{1}{c}{\multirow{4}[2]{*}{2}} & 1     & 0.9   & 0.8   & 0.7   & 0.6   &       & \multicolumn{1}{c}{\multirow{4}[2]{*}{2}} & 1     & 0.9   & 0.8   & 0.7   & 0.6   &       & \multicolumn{1}{c}{\multirow{4}[2]{*}{2}} & 1     & 0.9   & 0.9   & 0.9   & 0.8   &       & \multicolumn{1}{c}{\multirow{4}[2]{*}{2}} & 1     & 0.9   & 0.9   & 0.9   & 0.8 \\
		\multicolumn{1}{c}{} & 2     & 0.1   & 0.2   & 0.3   & 0.4   &       & \multicolumn{1}{c}{} & 2     & 0.1   & 0.2   & 0.3   & 0.4   &       & \multicolumn{1}{c}{} & 2     & 0.1   & 0.1   & 0.1   & 0.2   &       & \multicolumn{1}{c}{} & 2     & 0.1   & 0.1   & 0.1   & 0.2 \\
		\multicolumn{1}{c}{} & 3     & 0.1   & 0.2   & 0.3   & 0.4   &       & \multicolumn{1}{c}{} & 3     & 0.1   & 0.2   & 0.3   & 0.4   &       & \multicolumn{1}{c}{} & 3     & 0.1   & 0.1   & 0.1   & 0.2   &       & \multicolumn{1}{c}{} & 3     & 0.1   & 0.1   & 0.1   & 0.2 \\
		\multicolumn{1}{c}{} & 4     & 0.9   & 0.8   & 0.7   & 0.6   &       & \multicolumn{1}{c}{} & 4     & 0.9   & 0.8   & 0.7   & 0.6   &       & \multicolumn{1}{c}{} & 4     & 0.9   & 0.9   & 0.9   & 0.8   &       & \multicolumn{1}{c}{} & 4     & 0.9   & 0.9   & 0.9   & 0.8 \\
		\hline
		\multicolumn{1}{c}{\multirow{4}[2]{*}{3}} & 1     & 0.9   & 0.8   & 0.7   & 0.6   &       & \multicolumn{1}{c}{\multirow{4}[2]{*}{3}} & 1     & 0.9   & 0.8   & 0.7   & 0.6   &       & \multicolumn{1}{c}{\multirow{4}[2]{*}{3}} & 1     & 0.9   & 0.8   & 0.7   & 0.7   &       & \multicolumn{1}{c}{\multirow{4}[2]{*}{3}} & 1     & 0.9   & 0.9   & 0.9   & 0.8 \\
		\multicolumn{1}{c}{} & 2     & 0.1   & 0.2   & 0.3   & 0.4   &       & \multicolumn{1}{c}{} & 2     & 0.1   & 0.2   & 0.3   & 0.4   &       & \multicolumn{1}{c}{} & 2     & 0.1   & 0.2   & 0.3   & 0.3   &       & \multicolumn{1}{c}{} & 2     & 0.1   & 0.1   & 0.1   & 0.2 \\
		\multicolumn{1}{c}{} & 3     & 0.1   & 0.2   & 0.3   & 0.4   &       & \multicolumn{1}{c}{} & 3     & 0.1   & 0.2   & 0.3   & 0.4   &       & \multicolumn{1}{c}{} & 3     & 0.1   & 0.2   & 0.3   & 0.3   &       & \multicolumn{1}{c}{} & 3     & 0.1   & 0.1   & 0.1   & 0.2 \\
		\multicolumn{1}{c}{} & 4     & 0.9   & 0.8   & 0.7   & 0.6   &       & \multicolumn{1}{c}{} & 4     & 0.9   & 0.8   & 0.7   & 0.6   &       & \multicolumn{1}{c}{} & 4     & 0.9   & 0.8   & 0.7   & 0.7   &       & \multicolumn{1}{c}{} & 4     & 0.9   & 0.9   & 0.9   & 0.8 \\
		\hline
		\multicolumn{1}{c}{} &       &       &       &       &       &       & \multicolumn{1}{c}{\multirow{4}[2]{*}{4}} & 1     & 0.9   & 0.8   & 0.7   & 0.6   &       & \multicolumn{1}{c}{\multirow{4}[2]{*}{4}} & 1     & 0.9   & 0.8   & 0.7   & 0.7   &       & \multicolumn{1}{c}{\multirow{4}[2]{*}{4}} & 1     & 0.9   & 0.8   & 0.7   & 0.7 \\
		&       &       &       &       &       &       & \multicolumn{1}{c}{} & 2     & 0.1   & 0.2   & 0.3   & 0.4   &       & \multicolumn{1}{c}{} & 2     & 0.1   & 0.2   & 0.3   & 0.3   &       & \multicolumn{1}{c}{} & 2     & 0.1   & 0.2   & 0.3   & 0.3 \\
		&       &       &       &       &       &       & \multicolumn{1}{c}{} & 3     & 0.1   & 0.2   & 0.3   & 0.4   &       & \multicolumn{1}{c}{} & 3     & 0.1   & 0.2   & 0.3   & 0.3   &       & \multicolumn{1}{c}{} & 3     & 0.1   & 0.2   & 0.3   & 0.3 \\
		&       &       &       &       &       &       & \multicolumn{1}{c}{} & 4     & 0.9   & 0.8   & 0.7   & 0.6   &       & \multicolumn{1}{c}{} & 4     & 0.9   & 0.8   & 0.7   & 0.7   &       & \multicolumn{1}{c}{} & 4     & 0.9   & 0.8   & 0.7   & 0.7 \\
		\hline
		&       &       &       &       &       & \multicolumn{1}{c}{} &       &       &       &       &       &       &       & \multicolumn{1}{c}{\multirow{4}[2]{*}{5}} & 1     & 0.9   & 0.8   & 0.7   & 0.7   &       & \multicolumn{1}{c}{\multirow{4}[2]{*}{5}} & 1     & 0.9   & 0.8   & 0.7   & 0.6 \\
		&       &       &       &       &       & \multicolumn{1}{c}{} &       &       &       &       &       &       &       & \multicolumn{1}{c}{} & 2     & 0.1   & 0.2   & 0.3   & 0.3   &       & \multicolumn{1}{c}{} & 2     & 0.1   & 0.2   & 0.3   & 0.4 \\
		&       &       &       &       &       &       &       &       &       &       &       &       &       & \multicolumn{1}{c}{} & 3     & 0.1   & 0.2   & 0.3   & 0.3   &       & \multicolumn{1}{c}{} & 3     & 0.1   & 0.2   & 0.3   & 0.4 \\
		&       &       &       &       &       &       &       &       &       &       &       &       &       & \multicolumn{1}{c}{} & 4     & 0.9   & 0.8   & 0.7   & 0.7   &       & \multicolumn{1}{c}{} & 4     & 0.9   & 0.8   & 0.7   & 0.6 \\
		\hline
		&       &       &       &       &       &       &       &       &       &       &       &       &       &       &       &       &       &       &       &       & \multicolumn{1}{c}{\multirow{4}[2]{*}{6}} & 1     & 0.9   & 0.8   & 0.7   & 0.6 \\
		&       &       &       &       &       &       &       &       &       &       &       &       &       &       &       &       &       &       &       &       & \multicolumn{1}{c}{} & 2     & 0.1   & 0.2   & 0.3   & 0.4 \\
		&       &       &       &       &       &       &       &       &       &       &       &       &       &       &       &       &       &       &       &       & \multicolumn{1}{c}{} & 3     & 0.1   & 0.2   & 0.3   & 0.4 \\
		&       &       &       &       &       &       &       &       &       &       &       &       &       &       &       &       &       &       &       &       & \multicolumn{1}{c}{} & 4     & 0.9   & 0.8   & 0.7   & 0.6 \\
		\hline
	\end{tabular}%
	\label{lambdas}%
\end{table}
\begin{landscape}
	\renewcommand{\arraystretch}{2}
	\begin{table}[h]
		\centering
		\caption{Misclassification rate of $k$NN, tree, random forest, node harvest, SVM, OTE, $OTE_{oob}$ and $OTE_{sub}$. Number of observations is 1000 for each model. Bayes error is given in the fourth column against each model.}
		\setlength{\tabcolsep}{5pt}
		\begin{tabular}{lcccccccccccccccc}
			\toprule
			\multicolumn{1}{l}{Scenario} & \multicolumn{1}{l}{Number of } & \multicolumn{1}{l}{Number of } & Bayes Error & kNN   & Tree  & RF    & NH    & SVM   & SVM   & SVM   & SVM   & OTE   & OTE   & OTE   & $OTE_{oob}$ & $OTE_{sub}$ \\
			& \multicolumn{1}{l}{Variables} & \multicolumn{1}{l}{Observations} &       &       &       &       &       & (Radial) & (Linear) & (Bessel) & (Laplacian) & 10\% V & 20\% V & 30\% V &       &  \\
			\midrule
			&       &       & 9.2\% & 23\% & 11\% & 10\% & 11\% & 20\% & 20\% & 20\% & 20\% & 10\% & 11\% & 12\% & 10\% & 11\% \\
			&       &       & 14\% & 27\% & 16\% & 15\% & 16\% & 23\% & 23\% & 24\% & 22\% & 15\% & 16\% & 17\% & 15\% & 17\% \\
			\multicolumn{1}{l}{Scenario 1} & 9     & 1000  & 18\% & 33\% & 19\% & 23\% & 26\% & 29\% & 29\% & 29\% & 29\% & 24\% & 25\% & 27\% & 24\% & 26\% \\
			&       &       & 33\% & 43\% & 37\% & 36\% & 38\% & 39\% & 39\% & 39\% & 38\% & 37\% & 39\% & 40\% & 37\% & 39\% \\
			\midrule
			&       &       & 21\% & 31\% & 23\% & 21\% & 23\% & 25\% & 24\% & 30\% & 25\% & 21\% & 23\% & 24\% & 21\% & 21\% \\
			&       &       & 24\% & 31\% & 26\% & 24\% & 25\% & 27\% & 27\% & 33\% & 27\% & 24\% & 25\% & 26\% & 24\% & 24\% \\
			\multicolumn{1}{l}{Scenario 2} & 12    & 1000  & 28\% & 37\% & 31\% & 29\% & 30\% & 32\% & 31\% & 37\% & 32\% & 29\% & 30\% & 31\% & 29\% & 30\% \\
			&       &       & 30\% & 40\% & 33\% & 32\% & 33\% & 34\% & 34\% & 38\% & 33\% & 32\% & 32\% & 33\% & 32\% & 32\% \\
			\midrule
			&       &       & 16\% & 32\% & 23\% & 19\% & 23\% & 25\% & 25\% & 56\% & 25\% & 20\% & 21\% & 22\% & 19\% & 19\% \\
			&       &       & 18\% & 34\% & 25\% & 23\% & 25\% & 27\% & 26\% & 56\% & 27\% & 23\% & 24\% & 25\% & 23\% & 23\% \\
			\multicolumn{1}{l}{Scenario 3} & 15    & 1000  & 21\% & 34\% & 26\% & 25\% & 28\% & 28\% & 28\% & 56\% & 28\% & 25\% & 27\% & 28\% & 25\% & 25\% \\
			&       &       & 24\% & 37\% & 31\% & 28\% & 30\% & 30\% & 30\% & 57\% & 31\% & 28\% & 30\% & 32\% & 28\% & 28\% \\
			\midrule
			&       &       & 22\% & 35\% & 29\% & 23\% & 26\% & 26\% & 26\% & 72\% & 28\% & 24\% & 25\% & 26\% & 23\% & 24\% \\
			&       &       & 22\% & 36\% & 29\% & 24\% & 27\% & 29\% & 28\% & 72\% & 28\% & 25\% & 27\% & 28\% & 24\% & 25\% \\
			\multicolumn{1}{l}{Scenario 4} & 18    & 1000  & 25\% & 40\% & 32\% & 26\% & 30\% & 32\% & 33\% & 68\% & 36\% & 30\% & 31\% & 32\% & 30\% & 30\% \\
			&       &       & 27\% & 41\% & 32\% & 29\% & 32\% & 33\% & 34\% & 70\% & 37\% & 31\% & 33\% & 34\% & 31\% & 32\% \\
			\bottomrule
		\end{tabular}%
		\label{sim1000}%
	\end{table}%
\end{landscape}
\begin{landscape}
	\begin{table}[htbp]
		\caption{Misclassification rate of $k$NN, tree, random forest, node harvest, SVM, OTE, $OTE_{oob}$ and $OTE_{sub}$. Number of observations is 100 for each model. Bayes error is given in the fourth column against each model.}
		\setlength{\tabcolsep}{4pt}
		\begin{tabular}{lccccccccccccc}
			\toprule
			\multicolumn{1}{l}{Scenario} & \multicolumn{1}{l}{Number of } & \multicolumn{1}{l}{Number of } & kNN   & Tree  & RF    & NH    & SVM   & SVM   & SVM   & SVM   & OTE   & $OTE_{oob}$ & $OTE_{sub}$ \\
			& \multicolumn{1}{l}{Variables} & \multicolumn{1}{l}{Observations} &       &       &       &       & (Radial) & (Linear) & (Bessel) & (Laplacian) &       &       &  \\
			\midrule
			&       &       & 29\% & 25\% & 23\% & 24\% & 25\% & 23\% & 28\% & 25\% & 27\% & 21\% & 20\% \\
			&       &       & 32\% & 28\% & 27\% & 27\% & 30\% & 28\% & 37\% & 30\% & 30\% & 26\% & 25\% \\
			\multicolumn{1}{l}{Scenario 1} & 9     & 100   & 36\% & 33\% & 31\% & 34\% & 34\% & 33\% & 37\% & 34\% & 34\% & 33\% & 32\% \\
			&       &       & 39\% & 46\% & 43\% & 45\% & 43\% & 42\% & 44\% & 44\% & 42\% & 42\% & 43\% \\
			\midrule
			&       &       & 37\% & 29\% & 28\% & 29\% & 30\% & 29\% & 38\% & 30\% & 31\% & 29\% & 27\% \\
			&       &       & 39\% & 32\% & 30\% & 32\% & 32\% & 31\% & 40\% & 33\% & 34\% & 32\% & 30\% \\
			\multicolumn{1}{l}{Scenario 2} & 12    & 100   & 40\% & 38\% & 36\% & 38\% & 37\% & 36\% & 42\% & 39\% & 38\% & 37\% & 37\% \\
			&       &       & 41\% & 41\% & 38\% & 41\% & 40\% & 38\% & 45\% & 42\% & 39\% & 40\% & 40\% \\
			\midrule
			&       &       & 39\% & 35\% & 31\% & 32\% & 31\% & 30\% & 53\% & 34\% & 33\% & 32\% & 30\% \\
			&       &       & 40\% & 36\% & 32\% & 34\% & 32\% & 32\% & 53\% & 36\% & 34\% & 33\% & 32\% \\
			\multicolumn{1}{l}{Scenario 3} & 15    & 100   & 42\% & 37\% & 31\% & 36\% & 34\% & 33\% & 53\% & 37\% & 34\% & 33\% & 33\% \\
			&       &       & 46\% & 40\% & 35\% & 39\% & 36\% & 36\% & 52\% & 40\% & 37\% & 36\% & 37\% \\
			\midrule
			&       &       & 40\% & 36\% & 32\% & 34\% & 33\% & 33\% & 63\% & 40\% & 36\% & 34\% & 32\% \\
			&       &       & 42\% & 37\% & 33\% & 37\% & 34\% & 34\% & 62\% & 40\% & 36\% & 35\% & 34\% \\
			\multicolumn{1}{l}{Scenario 4} & 18    & 100   & 63\% & 38\% & 36\% & 39\% & 35\% & 35\% & 63\% & 42\% & 39\% & 38\% & 37\% \\
			&       &       & 46\% & 39\% & 38\% & 40\% & 38\% & 37\% & 63\% & 44\% & 40\% & 41\% & 40\% \\
			\bottomrule
		\end{tabular}%
		\label{sim100}%
	\end{table}%
\end{landscape}
\twocolumn
Arbitrary constants $\theta_1$ and $\theta_2$ are taken as 0.5 and 15, respectively,  for all cases (models and scenarios). To see how the methods perform in small and relatively large sample situations, first we consider generating a total of $n = 1000$ observation using the above setup. All the methods; $k$NN, CART, node harvest, random forest, , SVM (with four different kernels), OTE, OTE$_{oob}$ and OTE$_{sub}$ are trained by using 70\% of the available data as training data and the remaining 30\% of the data is used as test data. For OTE$_{sub}$, random sample without replacement for growing the trees are taken from 90\% of the training data and the remaining 10\% of the training data are used for trees assessment based on individual and ensemble performance.  A total of $T=1000$ trees are grown for OTE as the initial ensemble. For all the methods considered, the same training and test parts are used. Under each scenario, experiments are iterated 1000 times thus getting 1000 realizations of the data for all the methods. Averaging results under the 1000 realizations gives the final results in all the cases. The results are given in Tables \ref{sim1000} and \ref{sim100}. Node weights $\lambda_{ijk}$ are altered in a way that lead to patterns in the data less or more meaningful and thus getting a high or low Bayes error as shown in column 4 of Table \ref{sim1000}. For each of the scenario, four different values of the Bayes error are obtained. The simulation also show that Bayes error of a simulation scenario can be changed by altering the number of trees and/or the node weights. For instance, weights of $0.1$ and $0.9$ given to extreme nodes (left most and right most) and internal nodes, respectively, will lead to a tree that is less complex compared to a tree with $0.2$ and $0.8$ such weights. For further explanation, see \cite{khan2016ensemble}. 

Unsurprisingly, tree and $k$NN have the maximum errors in all the cases of the four scenarios. \emph{OTE} and Random forest performed comparable with little variations in some cases. For OTE, three values of validation set size $V=10\%, 20\%, 30\%$ are considered. As can be seen in Table \ref{sim1000}, that increasing number of observations $V$ in the validation set the performance of OTE decrease in all the cases of each scenario.  In cases where the models generate data with meaningful patterns indicated by low Bayes errors, the results of OTE$_{oob}$  and random forest are better or comparable. OTE$_{sub}$ did not perform well compared to random forest, OTE and OTE$_{oob}$. The reason for this might be that as OTE$_{oob}$ selects only a few trees for the final ensemble, enough randomness in trees could not be guaranteed by growing them on samples of size 90\% of training data size drawn without replacement.  SVM show similar results to $k$NN and tree in almost all the cases.

To see how the methods perform in relatively small sample situations, the same simulation scenarios are used to generate datasets consisting of $n=100$ observations. The results are given in Table \ref{sim100}. This time OTE$_{sub}$ outperforms the rest of the methods in datasets with meaningful structures, i.e. with low Bayes error (not shown). A few accurate and diverse trees can better capture the patterns in the relatively small sized data compared to other methods. In datasets with less meaningful structures, the method still performs similar to SVM which is considered as a promissing classifier in small sample situations. The overall performance of SVM relative to other methods has also improved as compared to the previous situation with $n=1000$.
\subsection{Analysis of benchmark data sets}
This section presents our analysis on benchmark data sets for OTE$_{oob}$, OTE$_{sub}$ and the other methods considered. A total of 21 data sets are used for comparison purposes. These data sets are described briefly in Table \ref{datasets}. Against each dataset, the number $n$ of observations, number $d$ of features and the corresponding sources from where the data can be taken, are given. Number of features by feature type are also given against each data set. The domain of each dataset is also given in the table. 
\onecolumn
\renewcommand{\arraystretch}{1.5}
\begin{table}[h]
	\centering
	\caption{Datasets for regression and classification with the corresponding number of instances $n$, number of variable $d$ and feature/variable type; F: real, I: integer and N: nominal variable in a data set. Sources of the data and their domain are also given.}
	\setlength{\tabcolsep}{3pt}
	\fontsize{10}{10}\selectfont
	\begin{tabular}{lccccc}
		\toprule
		\textbf{Data Set} & $n$ & $d$ & \textbf{Feature type}& \textbf{Source}& Domain \\
		\textbf{} &  &  & \textbf{(R/I/N)} &&\\
		\midrule
		Mammographic & 830   & 5     & \multicolumn{1}{c}{(0/5/0)} & \fontsize{8}{8}\selectfont \url{http://sci2s.ugr.es/keel/category.php?cat=clas}& Medical Science\\
		Dystrophy & 209   & 5     & \multicolumn{1}{c}{(2/3/0)} & \cite{ipred}& Medical Science\\
		Monk3 & 122   & 6     & \multicolumn{1}{c}{(0/6/0)} & \cite{frank2013uci}&  Machine Learning Benchmark\\
		Appendicitis & 106   & 7     & \multicolumn{1}{c}{(7/0/0)}&\fontsize{8}{8}\selectfont\url{http://sci2s.ugr.es/keel/dataset.php?cod=183} & Medical Science\\
		SAHeart & 462   & 9     & \multicolumn{1}{c}{(5/3/1)} & \fontsize{8}{8}\selectfont\url{http://sci2s.ugr.es/keel/dataset.php?cod=184#sub1}&Medical Science\\
		Tic-Tac-Toe & 958   & 9     & \multicolumn{1}{c}{(0/0/9)}& \cite{frank2013uci}& Gambling\\
		Heart & 303   & 13    & \multicolumn{1}{c}{(1/12/0)} &\cite{frank2013uci}& Medical Science\\
		House vote & 232   & 16    & \multicolumn{1}{c}{(0/0/16)}& \cite{frank2013uci}&Medical Science\\
		Bands & 365   & 19    & \multicolumn{1}{c}{ (13/6/0)} &\fontsize{8}{8}\selectfont \url{http://sci2s.ugr.es/keel/dataset.php?cod=184#sub1}& Physical Science\\
		Hepatitis & 80    & 20    & \multicolumn{1}{c}{(2/18/0)} &  \cite{frank2013uci} &Medical Science\\
		Parkinson & 195   & 22    & \multicolumn{1}{c}{(22/0/0)} &  \cite{frank2013uci}&Medical Science\\
		Body  & 507   & 23    & \multicolumn{1}{c}{(22/1/0)} & \cite{gclus}&Medical Science\\
		Thyroid & 9172  & 27    & \multicolumn{1}{c}{(3/2/22)}& \cite{frank2013uci}&Medical Science\\
		WDBC  & 569   & 29    & \multicolumn{1}{c}{(29/0/0)} &\cite{frank2013uci}&Medical Science\\
		WPBC  & 198   & 32    & \multicolumn{1}{c}{(30/2/0)} &\cite{frank2013uci}&Medical Science\\
		Oil-Spill & 937   & 49    & \multicolumn{1}{c}{(40/9/0)} &\url{http://openml.org/}& Environmental Science\\
		Spam base & 4601  & 57    & \multicolumn{1}{c}{(55/2/0)}& \cite{frank2013uci}&Fraud Detection\\
		Glaucoma & 196   & 62    & \multicolumn{1}{c}{(62/0/0)}& \cite{ipred}&Medical Science\\
		Nki 70 & 144   & 76    & \multicolumn{1}{c}{(71/5/0)} &\cite{penalized}&Medical Science\\
		Musk  & 476   & 166   & \multicolumn{1}{c}{(0/166/0)} &\cite{kernlab}&Chemical Science\\
		\bottomrule
	\end{tabular}%
	\label{datasets}
\end{table}%
\twocolumn
\subsection{Experimental Setup} \label{setup}
Experimental setup for applying the methods on the 21 datasets is as follows. Given datassets are divided into training and testing parts consisting of 70\%  and 30\%, respectively, of the total data. Splitting into $50\% - 50\%$  and $30\% - 70\%$ parts for training-testing are also considered. A total of 1000 random splittings of the given data are done into training and testing parts with methods trained on the training parts and tested by testing parts. Final result is obtained by averaging the results of all these 1000 splittings.

For  the original OTE, various values of validation set sizes, i.e. $|V|= 10, 15, 20$, are used to see its effect on the predictive performance of the method. A total of $T=1500$ trees are grown on independent bootstrap samples from the respective 90\% , 85\% and 80\% of training data by the method of random forest. The remaining 10\%, 15\% and 20\% data, respectively, are used for internal validation as mentioned above. 

For OTE$_{oob}$, $T = 1500$ trees are grown on bootstrap samples from the whole of the available training data. OOB observations are stored for individual and ensemble tree assessment. For OTE$_{sub}$ $T=1500$ trees are grown on random samples without replacement of size 90\% of training data size. The remaining 10\% are used for individual and ensemble performance assessment of each corresponding tree. The number $p$ of features is fixed at $p=\sqrt{d}$ for all data sets.  $M$ is fixed at 20\% of $T$. 

Various hyper-parameters of CART are tuned by using the \texttt{tune.rpart} R-function available within the R-Package \texttt{e1071} \cite{e1071}. Various values, (5,10,15,20,25) are tried to find the minimal optimal depth and optimal number of splits for the trees.

In the case of random forest, node size (\texttt{nodesize}), number of trees (\texttt{ntree}) and subset size ($p$) of features (\texttt{mtry}) are tuned by using \texttt{tune.randomForest} function available with in the R-Package \texttt{e1071} as used by \cite{khan2016ensemble,adler2008comparison}. Searches for the best node size (\texttt{nodesize}) are mede among values (1,5,10,15,20,25,30), for \texttt{ntree} amongst values (500,1000,1500,2000) and for \texttt{mtry} (sqrt(d), d/5, d/4, d/3, d/2) are checked. All the possible values of \texttt{mtry} are checked where $d < 12$.

For node harvest estimator, the only heper-parameter is the number of nodes in the initial ensemble.  Meinshausen \cite{meinshausen2010node} has shown that for its large values the changes in the results are negligable and stated that initial number of nodes greater than 1000 gives almost the same results. In this paper, this value is fixed at 1500. R implementation as given in the package \texttt{nodeHarvest} \cite{nodeharvest} is used. For support vector machine, automatic estimation of sigma is utalised from the R package \texttt{kernlab} \cite{kernlab}. For the remining parameters, their default values are used with four kernels, Radial, Linear, Bessel and Laplacian. $k$-nearest neighbours classifier, $k$NN, is tuned for the optimal value of its hyper-parameter $k$, the number of nearest neighbouts, by using \texttt{tune.knn} R function within the R library \texttt{e1071}. Values of $k = 1, \ldots, 10$ are tried.

For random projection (RP) ensemble method\cite{RPEnsemble2017paper}, the R package \texttt{RPEnsemble} \cite{RPEnsemble} is used. Due to computational constraint $B_1$ and $B_2$ are fixed at 30 and 5 respectively. Quadratic discriminant analysis \texttt{base = "QDA"} and linear discriminant analysis \texttt{base = "LDA"} procedures are used as the base learner along with \texttt{d=5, projmethod = "Haar"}. The remaining parameters are kept at their default values.

For a fair comparison, training and test data are taken the same for tree, node harvest, random forest, SVM, RP, OTE, OTE$_{oob}$ and OTE$_{oob}$. Average classification errors are recorded for all the methods on all the data sets. R version 4.0.1 \cite{R}, on a 3 GHz Intel Core i7 computer with 8 GB memory running under mac OS X operating system, is used for the experiments. The results for various training and testing parts are given in Tables \ref{results70}, \ref{results50} and \ref{results30}. For further assessment of the proposed methods in comparison with the rest, Brier score, sensitivity and Kappa statistics values are also estimated. These statistics are estimated based on 30\% training and 70\% testing partitions of the given datasets for checking the behaviour of the ensembles in small sample training data. The results in terms of Brier score, sensitivity and Kappa are given in Tables \ref{brier}, \ref{sen} and \ref{kappa}, respectively.

\begin{landscape}
	\renewcommand{\arraystretch}{2}
	\begin{table}[h]
			\caption{Misclassification rates of $k$NN, tree, random forest, node harvest, SVM (with four kernels), random projection with linear and quadratic discriminant analyses, OTE, OTE$_{oob}$ and OTE$_{sub}$. Results are based on 70\% training and 30\% testing parts of the data. Overall best performing method result is shown in bold. The results are italicised when OTE$_{oob}$ and/or OTE$_{sub}$ are/is better than \emph{OTE}.}
			\label{results70}%
			\setlength{\tabcolsep}{4pt}
			\begin{tabular}{lccccccccccccccccc}
				\toprule
				Dataset & \multicolumn{1}{l}{$n$} & \multicolumn{1}{l}{$p$} & \multicolumn{1}{l}{$k$NN} & \multicolumn{1}{l}{Tree} & \multicolumn{1}{c}{NH} & \multicolumn{1}{c}{SVM} & \multicolumn{1}{c}{SVM} & \multicolumn{1}{c}{SVM} & \multicolumn{1}{c}{SVM} & \multicolumn{1}{c}{RP} & \multicolumn{1}{c}{RP} & \multicolumn{1}{c}{RF} &       & \multicolumn{1}{c}{OTE} &       & \multicolumn{1}{c}{OTE$_{oob}$} & \multicolumn{1}{c}{OTE$_{sub}$} \\
				&       &       &       &       &       & \multicolumn{1}{c}{(Radial)} & \multicolumn{1}{c}{(Linear)} & \multicolumn{1}{c}{(Bessel)} & \multicolumn{1}{c}{(Laplacian)} & \multicolumn{1}{c}{(LDA)} & \multicolumn{1}{c}{(QDA)} &       & \multicolumn{1}{c}{10\% V} & \multicolumn{1}{c}{15\% V} & \multicolumn{1}{c}{20\% V} &       &  \\
				\midrule
				Mammographic & 830   & 5     & 0.2015 & 0.1648 & \textbf{0.1632} & 0.1882 & 0.1763 & 0.1862 & 0.1860 & 0.1928 & 0.1966 & 0.1643 & 0.1808 & 0.1804 & 0.1801 & 0.1839 & 0.1912 \\
				Dystrophy & 209   & 5     & 0.1294 & 0.1495 & 0.1527 & 0.1039 & 0.1137 & 0.1085 & 0.1038 & 0.1221 & \textbf{0.0946} & 0.1256 & 0.1226 & 0.1247 & 0.1263 & \textit{0.1218} & 0.1243 \\
				Monk3 & 122   & 6     & 0.1388 & 0.0946 & 0.2740 & 0.1048 & 0.2325 & 0.1007 & 0.1380 & 0.2177 & 0.1150 & 0.0708 & 0.0802 & 0.0761 & 0.0755 & \textit{\textbf{0.0701}} & 0.0780 \\
				Appendicitis & 106   & 7     & 0.1511 & 0.1569 & 0.1458 & 0.2086 & 0.1771 & 0.1883 & 0.1599 & \textbf{0.1280} & 0.1559 & 0.1336 & 0.1519 & 0.1534 & 0.1547 & \textit{0.1451} & 0.1470 \\
				SAHeart & 462   & 9     & 0.3471 & 0.3241 & \textbf{0.2826} & 0.3132 & 0.3134 & 0.3403 & 0.3191 & 0.3036 & 0.3057 & 0.2995 & 0.3184 & 0.3192 & 0.3190 & \textit{0.3166} & 0.3286 \\
				Tic-Tac-Toe & 958   & 9     & 0.3798 & 0.1955 & 0.2916 & 0.2392 & 0.4013 & 0.1943 & 0.3800 & 0.3151 & 0.2424 & 0.0511 & 0.0557 & 0.0649 & 0.0693 & 0.0579 & \textbf{0.0421} \\
				Heart & 303   & 13    & 0.3608 & 0.2290 & 0.1999 & 0.1734 & 0.1779 & 0.3865 & \textbf{0.1672} & 0.2033 & 0.2159 & 0.1738 & 0.1995 & 0.2023 & 0.2031 & \textit{0.1845} & 0.1875 \\
				House Vote & 232   & 16    & 0.0911 & 0.0443 & 0.1112 & 0.0486 & 0.0456 & 0.0477 & 0.0418 & 0.0686 & 0.0687 & \textbf{0.0401} & 0.0423 & 0.0429 & 0.0435 & \textit{0.0420} & 0.0416 \\
				Bands & 365   & 19    & 0.3255 & 0.3056 & 0.3775 & 0.2988 & 0.2798 & 0.3622 & 0.5279 & 0.3379 & 0.3126 & 0.2333 & 0.2379 & 0.2445 & 0.2482 & 0.2380 & \textbf{0.2301} \\
				Hepatitis & 80    & 20    & 0.3966 & 0.1846 & 0.1330 & 0.1368 & 0.1650 & 0.4884 & 0.1576 & 0.1883 & 0.1512 & 0.1530 & 0.1320 & 0.1353 & 0.1370 & \textit{0.1305} & \textbf{0.0419} \\
				Parkinson & 195   & 22    & 0.1752 & 0.1414 & 0.1334 & 0.1622 & 0.2012 & 0.2653 & 0.2089 & 0.1812 & 0.1568 & 0.1026 & 0.0976 & 0.0994 & 0.1011 & \textit{0.0966} & \textbf{0.0951} \\
				Body  & 507   & 23    & 0.0315 & 0.0828 & 0.0854 & 0.0186 & 0.0170 & 0.5419 & \textbf{0.0352} & 0.0208 & 0.0242 & 0.0435 & 0.0409 & 0.0429 & 0.0438 & 0.0410 & 0.0382 \\
				Thyroid & 9172  & 27    & 0.0391 & 0.0128 & 0.0290 & 0.1118 & 0.0328 & 0.3084 & 0.0766 & 0.0497 & 0.0440 & \textbf{0.0102} & 0.0105 & 0.0107 & 0.0108 & \textit{0.0103} & \textbf{0.0102} \\
				WDBC  & 569   & 29    & 0.0771 & 0.0683 & 0.0613 & 0.0432 & 0.0272 & 0.6175 & 0.0444 & 0.0575 & 0.0564 & 0.0419 & 0.0416 & 0.0426 & 0.0438 & \textit{0.0404} & \textbf{0.0389} \\
				WPBC  & 198   & 32    & 0.2691 & 0.2953 & 0.2328 & 0.2960 & 0.2862 & 0.5521 & 0.3546 & 0.2198 & 0.2320 & 0.2088 & \textbf{0.2077} & 0.2139 & 0.2181 & 0.2078 & 0.2127 \\
				Oil-Spill & 937   & 49    & 0.0562 & 0.0394 & 0.0410 & 0.0774 & 0.0956 & 0.3641 & 0.1189 & 0.0440 & 0.0434 & 0.0371 & 0.0348 & 0.0354 & 0.0357 & \textit{0.0347} & \textbf{0.0344} \\
				Spam base & 4601  & 58    & 0.1788 & 0.1064 & 0.1004 & 0.0917 & 0.0743 & 0.4919 & 0.1058 & 0.2157 & 0.2258 & 0.0493 & 0.0493 & 0.0503 & 0.0486 & \textit{0.0482} & \textbf{0.0468} \\
				Sonar & 208   & 60    & 0.1819 & 0.2901 & 0.2429 & 0.1832 & 0.2562 & 0.5389 & 0.3011 & 0.2624 & 0.2138 & 0.1916 & 0.1823 & 0.1887 & 0.1925 & \textit{0.1746} & \textbf{0.1642} \\
				Glaucoma & 196   & 62    & 0.2010 & 0.1339 & 0.1254 & 0.1200 & 0.1573 & 0.6468 & 0.1233 & \textbf{0.1076} & 0.1432 & 0.1125 & 0.1125 & 0.1146 & 0.1166 & 0.1163 & 0.1190 \\
				Nki 70 & 144   & 76    & 0.1878 & 0.1662 & 0.1565 & 0.2278 & 0.3321 & 0.4031 & 0.4958 & 0.1813 & 0.1900 & 0.1460 & \textbf{0.1456} & 0.1478 & 0.1491 & 0.1494 & 0.1665 \\
				Musk  & 476   & 166   & 0.1438 & 0.2221 & 0.2547 & 0.1490 & 0.1623 & 0.4894 & 0.5187 & 0.1010 & \textbf{0.0844} & 0.1200 & 0.1134 & 0.1188 & 0.1248 & \textit{0.1079} & 0.1068 \\
				\bottomrule
			\end{tabular}%
	\end{table}%
\end{landscape}
\begin{landscape}
	\renewcommand{\arraystretch}{2}
	\begin{table}[h]
			\caption{Misclassification rates of $k$NN, tree, random forest, node harvest, SVM (with four different kernels), random projection with quadratic and linear discriminant analyses, OTE, OTE$_{oob}$ and OTE$_{sub}$. Results are based on 50\% training and 50\% testing parts of the data. Overall best performing method result is shown in bold. The results are italicised when OTE$_{oob}$ and/or OTE$_{sub}$ are/is better than \emph{OTE}.}
			\label{results50}%
			\setlength{\tabcolsep}{4pt}
			\begin{tabular}{lccccccccccccccc}
				\toprule
				Dataset & \multicolumn{1}{c}{$n$} & \multicolumn{1}{c}{$p$} & \multicolumn{1}{c}{kNN} & \multicolumn{1}{c}{Tree} & \multicolumn{1}{c}{NH} & \multicolumn{1}{c}{SVM} & \multicolumn{1}{c}{SVM} & \multicolumn{1}{c}{SVM} & \multicolumn{1}{c}{SVM} & \multicolumn{1}{c}{RP} & \multicolumn{1}{c}{RP} & \multicolumn{1}{c}{RF} & \multicolumn{1}{c}{OTE} & \multicolumn{1}{c}{OTE$_{oob}$} & \multicolumn{1}{c}{OTE$_{sub}$} \\
				&       &       &       &       &       & \multicolumn{1}{c}{(Radial)} & \multicolumn{1}{c}{(Linear)} & \multicolumn{1}{c}{(Bessel)} & \multicolumn{1}{c}{(Laplacian)} & \multicolumn{1}{c}{(LDA)} & \multicolumn{1}{c}{(QDA)} &       &       &       &  \\
				\midrule
				Mammographic & 830   & 5     & 0.2100 & 0.1661 & 0.1760 & 0.1882 & 0.1772 & 0.1836 & 0.1845 & 0.1890 & 0.1993 & \textbf{0.1614} & 0.1793 & 0.1824 & 0.1916 \\
				Dystrophy & 209   & 5     & 0.1354 & 0.1831 & 0.1632 & 0.1101 & 0.1191 & 0.1112 & 0.1087 & 0.1233 & \textbf{0.0959} & 0.1316 & 0.1311 & \textit{0.1285} & 0.1308 \\
				Monk3 & 122   & 6     & 0.1421 & 0.0773 & 0.2855 & 0.1228 & 0.2367 & 0.1137 & 0.1731 & 0.2308 & 0.1352 & 0.0790 & 0.0887 & \textit{\textbf{0.0735}} & 0.0833 \\
				Appendicitis & 106   & 7     & 0.1624 & 0.1777 & 0.1569 & 0.1937 & 0.2026 & 0.1885 & 0.1631 & \textbf{0.1358} & 0.1692 & 0.1418 & 0.1681 & \textit{0.1493} & 0.1526 \\
				SAHeart & 462   & 9     & 0.3499 & 0.3409 & \textbf{0.2876} & 0.3179 & 0.3153 & 0.3430 & 0.3249 & 0.3057 & 0.3140 & 0.2979 & 0.3184 & 0.3202 & 0.3282 \\
				Tic-Tac-Toe & 958   & 9     & 0.3790 & 0.1884 & 0.3109 & 0.2771 & 0.4073 & 0.2233 & 0.4462 & 0.3266 & 0.2536 & 0.0751 & 0.0791 & 0.0826 & \textit{\textbf{0.0642}} \\
				Heart & 303   & 13    & 0.3701 & 0.2370 & 0.2140 & 0.1761 & 0.1830 & 0.3807 & \textbf{0.1703} & 0.2120 & 0.2288 & 0.1790 & 0.2037 & \textit{0.1899} & 0.1929 \\
				House Vote & 232   & 16    & 0.0987 & 0.0472 & 0.1211 & 0.0514 & 0.0467 & 0.0497 & 0.0655 & 0.0699 & 0.0717 & \textbf{0.0418} & 0.0441 & \textit{0.0434} & 0.0438 \\
				Bands & 365   & 19    & 0.3299 & 0.3289 & 0.3827 & 0.3432 & 0.2969 & 0.4397 & 0.5185 & 0.3413 & 0.3184 & \textbf{0.2522} & 0.2580 & \textit{0.2557} & \textit{0.2525} \\
				Parkinson & 195   & 22    & 0.1822 & 0.1720 & 0.1430 & 0.1831 & 0.2071 & 0.2564 & 0.2234 & 0.1842 & 0.1671 & 0.1200 & 0.1143 & \textit{0.1135} & \textit{\textbf{0.1093}} \\
				Body  & 507   & 23    & 0.0410 & 0.1006 & 0.0887 & 0.0229 & 0.0187 & 0.5273 & 0.0420 & \textbf{0.0214} & 0.0240 & 0.0496 & 0.0473 & \textit{0.0472} & \textit{0.0454} \\
				Thyroid & 9172  & 27    & 0.0401 & 0.0130 & 0.0302 & 0.1153 & 0.0370 & 0.3546 & 0.0718 & 0.0500 & 0.0454 & 0.0110 & 0.0111 & \textit{0.0110} & \textit{\textbf{0.0108}} \\
				WDBC  & 569   & 29    & 0.0790 & 0.0749 & 0.0754 & 0.0503 & 0.3020 & 0.5979 & 0.0524 & 0.0584 & 0.0567 & 0.0461 & 0.0468 & \textit{0.0445} & \textit{\textbf{0.0443}} \\
				WPBC  & 198   & 32    & 0.2712 & 0.2836 & 0.2483 & 0.3117 & 0.2970 & 0.5208 & 0.4181 & \textbf{0.2242} & 0.2463 & 0.2246 & 0.2276 & \textit{0.2260} & 0.2324 \\
				Oil-Spill & 937   & 49    & 0.0654 & 0.0465 & 0.0521 & 0.0799 & 0.0933 & 0.3087 & 0.1273 & 0.0446 & 0.0435 & 0.0490 & 0.0380 & \textit{\textbf{0.0379}} & \textit{\textbf{0.0379}} \\
				Spam base & 4601  & 58    & 0.1912 & 0.1059 & 0.1124 & 0.0941 & 0.0758 & 0.4895 & 0.1190 & 0.2180 & 0.3149 & 0.0527 & 0.0522 & \textit{0.0520} & \textit{\textbf{0.0506}} \\
				Sonar & 208   & 60    & 0.1900 & 0.2984 & 0.2568 & 0.2045 & 0.2596 & 0.5393 & 0.4000 & 0.2689 & 0.2249 & 0.2075 & 0.2046 & \textit{0.2035} & \textit{\textbf{0.1920}} \\
				Glaucoma & 196   & 62    & 0.2102 & 0.1526 & 0.1375 & 0.1374 & 0.1663 & 0.6430 & 0.2024 & \textbf{0.1182} & 0.1439 & 0.1200 & 0.1279 & \textit{0.1255} & 0.1279 \\
				Nki 70 & 144   & 76    & 0.1912 & 0.1508 & 0.1648 & 0.2000 & 0.3281 & 0.3780 & 0.5097 & 0.1922 & 0.1984 & 0.1561 & \textbf{0.1545} & 0.1635 & 0.1865 \\
				Musk  & 476   & 166   & 0.1599 & 0.2583 & 0.2654 & 0.1700 & 0.1896 & 0.4946 & 0.5152 & 0.1128 & \textbf{0.0964} & 0.1415 & 0.1412 & \textit{0.1373} & \textit{0.1341} \\
				\bottomrule
			\end{tabular}%
	\end{table}%
\end{landscape}

\begin{landscape}
	\renewcommand{\arraystretch}{2}
	\begin{table}[h]
			\caption{Misclassification rates of $k$NN, tree, random forest, node harvest, SVM (with four different kernels), random projection with quadratic and linear discriminant analyses, OTE, OTE$_{oob}$ and OTE$_{sub}$. Results are based on 30\% training and 70\% testing parts of the data. Overall best performing method result is shown in bold. The results are italicised when OTE$_{oob}$ and/or OTE$_{sub}$ are/is better than \emph{OTE}.}
			\label{results30} %
			\setlength{\tabcolsep}{4pt}
			\begin{tabular}{lccccccccccccccc}
				\toprule
				Dataset & \multicolumn{1}{c}{$n$} & \multicolumn{1}{c}{$p$} & \multicolumn{1}{c}{kNN} & \multicolumn{1}{c}{Tree} & \multicolumn{1}{c}{NH} & \multicolumn{1}{c}{SVM} & \multicolumn{1}{c}{SVM} & \multicolumn{1}{c}{SVM} & \multicolumn{1}{c}{SVM} & \multicolumn{1}{c}{RP} & \multicolumn{1}{c}{RP} & \multicolumn{1}{c}{RF} & \multicolumn{1}{c}{OTE} & \multicolumn{1}{c}{OTE$_{oob}$} & \multicolumn{1}{c}{OTE$_{sub}$} \\
				&       &       &       &       &       & \multicolumn{1}{c}{(Radial)} & \multicolumn{1}{c}{(Linear)} & \multicolumn{1}{c}{(Bessel)} & \multicolumn{1}{c}{(Laplacian)} & \multicolumn{1}{c}{(LDA)} & \multicolumn{1}{c}{(QDA)} &       &       &       &  \\
				\midrule
				Mammographic & 830   & 5     & 0.2190 & 0.1665 & 0.1792 & 0.1911 & 0.1806 & 0.1836 & 0.1860 & 0.1857 & 0.1980 & \textbf{0.1640} & 0.1787 & 0.1827 & 0.1947 \\
				Dystrophy & 209   & 5     & 0.1399 & 0.2098 & 0.1725 & 0.1212 & 0.1272 & 0.1175 & 0.1176 & 0.1276 & \textbf{0.1041} & 0.1379 & 0.1458 & \textit{0.1392} & 0.1479 \\
				Monk3 & 122   & 6     & 0.1499 & 0.0883 & 0.2988 & 0.1732 & 0.2542 & 0.1610 & 0.2357 & 0.2466 & 0.1938 & 0.0973 & 0.1230 & \textit{\textbf{0.0874}} & 0.0962 \\
				Appendicitis & 106   & 7     & 0.1711 & 0.1998 & 0.1659 & 0.1954 & 0.2285 & 0.1889 & 0.1664 & 0.1506 & 0.1751 & \textbf{0.1496} & 0.1934 & \textit{0.1744} & \textit{0.1730} \\
				SAHeart & 462   & 9     & 0.3929 & 0.3549 & \textbf{0.2987} & 0.3299 & 0.3210 & 0.3500 & 0.3422 & 0.3108 & 0.3225 & 0.3044 & 0.3232 & 0.3258 & 0.3328 \\
				Tic-Tac-Toe & 958   & 9     & 0.3817 & 0.1988 & 0.3337 & 0.3453 & 0.4172 & 0.2783 & 0.5037 & 0.3381 & 0.2633 & 0.1343 & 0.1203 & \textit{\textbf{0.0828}} & 0.1067 \\
				Heart & 303   & 13    & 0.3798 & 0.2749 & 0.2259 & 0.1835 & 0.1999 & 0.3677 & \textbf{0.1838} & 0.2225 & 0.2456 & 0.1939 & 0.2155 & \textit{0.2029} & 0.2034 \\
				House Vote & 232   & 16    & 0.0995 & 0.0520 & 0.1357 & 0.0596 & 0.0483 & 0.0567 & 0.0810 & 0.0745 & 0.0783 & \textbf{0.0465} & 0.0509 & \textit{0.0471} & 0.0472 \\
				Bands & 365   & 19    & 0.3332 & 0.3433 & 0.3876 & 0.4465 & 0.3210 & 0.4970 & 0.5070 & 0.3462 & 0.3318 & 0.2780 & 0.2834 & \textit{\textbf{0.2771}} & 0.2815 \\
				Parkinson & 195   & 22    & 0.1899 & 0.1883 & 0.1687 & 0.2046 & 0.2150 & 0.2530 & 0.2370 & 0.1919 & 0.1891 & 0.1466 & 0.1433 & \textit{0.1401} & \textit{\textbf{0.1381}} \\
				Body  & 507   & 23    & 0.0419 & 0.0994 & 0.0901 & 0.0290 & 0.0216 & 0.5019 & 0.0555 & \textbf{0.0236} & 0.0259 & 0.0589 & 0.0579 & \textit{0.0563} & \textit{0.0540} \\
				Thyroid & 9172  & 27    & 0.0411 & 0.0137 & 0.0402 & 0.1289 & 0.0456 & 0.3479 & 0.0712 & 0.0501 & 0.0479 & 0.0131 & 0.0124 & \textit{0.0122} & \textit{\textbf{0.0120}} \\
				WDBC  & 569   & 29    & 0.0793 & 0.0802 & 0.0812 & 0.0604 & 0.0350 & 0.5698 & 0.0600 & 0.0602 & 0.0575 & 0.0537 & 0.0544 & \textit{0.0525} & \textit{\textbf{0.0513}} \\
				WPBC  & 198   & 32    & 0.2914 & 0.2985 & 0.2549 & 0.3273 & 0.3104 & 0.4366 & 0.4534 & 0.2348 & 0.2530 & \textbf{0.2446} & 0.2580 & \textit{0.2570} & 0.2624 \\
				Oil-Spill & 937   & 49    & 0.0697 & 0.0483 & 0.0621 & 0.0845 & 0.0824 & 0.1789 & 0.1754 & 0.0463 & 0.0429 & 0.0431 & 0.0411 & 0.0412 & \textit{\textbf{0.0409}} \\
				Spam base & 4601  & 58    & 0.1999 & 0.1078 & 0.1215 & 0.1176 & 0.0798 & 0.4829 & 0.3014 & 0.2368 & 0.3255 & 0.0577 & 0.0581 & \textit{0.0572} & \textit{\textbf{0.0564}} \\
				Sonar & 208   & 60    & 0.2562 & 0.3318 & 0.2659 & 0.2484 & 0.2714 & 0.5335 & 0.5051 & 0.2870 & 0.2605 & 0.2494 & 0.2453 & \textit{0.2378} & \textit{\textbf{0.2327}} \\
				Glaucoma & 196   & 62    & 0.2270 & 0.1953 & 0.1463 & 0.1681 & 0.1861 & 0.6269 & 0.2516 & 0.1365 & 0.1565 & 0.1589 & 0.1585 & \textit{\textbf{0.1519}} & \textit{0.1541} \\
				Nki 70 & 144   & 76    & 0.1979 & 0.1822 & 0.1758 & 0.2797 & 0.3304 & 0.3589 & 0.5099 & \textbf{0.1928} & 0.2103 & 0.1943 & 0.2218 & 0.2166 & 0.2310 \\
				Musk  & 476   & 166   & 0.1614 & 0.3058 & 0.2846 & 0.2147 & 0.2327 & 0.5015 & 0.5043 & 0.1280 & \textbf{0.1089} & 0.1849 & 0.1906 & \textit{0.1863} & \textit{0.1838} \\
				\bottomrule
			\end{tabular}%
	\end{table}%
\end{landscape}

\begin{landscape}
	\renewcommand{\arraystretch}{2}
	\begin{table}[h]
		\caption{Brier score of $k$NN, tree, random forest, node harvest, SVM (with four different kernels), random projection with quadratic and linear discriminant analyses, OTE, OTE$_{oob}$ and OTE$_{sub}$. Results are based on 30\% training and 70\% testing parts of the data. Overall best performing method result is shown in bold.}
		\label{brier} %
		\setlength{\tabcolsep}{4pt}
		\begin{tabular}{lccccccccccccccc}
			\toprule
			Dataset & \multicolumn{1}{c}{$n$} & \multicolumn{1}{c}{$p$} & \multicolumn{1}{c}{kNN} & \multicolumn{1}{c}{Tree} & \multicolumn{1}{c}{NH} & \multicolumn{1}{c}{SVM} & \multicolumn{1}{c}{SVM} & \multicolumn{1}{c}{SVM} & \multicolumn{1}{c}{SVM} & \multicolumn{1}{c}{RP} & \multicolumn{1}{c}{RP} & \multicolumn{1}{c}{RF} & \multicolumn{1}{c}{OTE} & \multicolumn{1}{c}{OTE$_{oob}$} & \multicolumn{1}{c}{OTE$_{sub}$} \\
			&       &       &       &       &       & \multicolumn{1}{c}{(Radial)} & \multicolumn{1}{c}{(Linear)} & \multicolumn{1}{c}{(Bessel)} & \multicolumn{1}{c}{(Laplacian)} & \multicolumn{1}{c}{(LDA)} & \multicolumn{1}{c}{(QDA)} &       &       &       &  \\
			\midrule
			Mammographic & 830  & 5   & 0.1633 & 0.1317 & \textbf{0.1255} & 0.1373 & 0.1354 & 0.1318 & 0.1338 & ... & ... & 0.1343 & 0.1398 & 0.1621 & 0.1529 \\
			Dystrophy    & 209  & 5   & 0.1191 & 0.1665 & 0.1167 & 0.1078 & \textbf{0.0929} & 0.0995 & 0.0936 & ... & ... & 0.0978 & 0.1060 & 0.1162 & 0.1086 \\
			Monk3        & 122  & 6   & 0.1401 & 0.0872 & 0.1425 & 0.1346 & 0.1886 & 0.1279 & 0.1524 & ... & ... & 0.0996 & 0.0813 & \textbf{0.0778} & 0.0802 \\
			Appendicitis & 106  & 7   & 0.1251 & 0.1504 & 0.1366 & 0.1385 & 0.1401 & 0.1323 & 0.1238 & ... & ... & \textbf{0.1200} & 0.1276 & 0.1487 & 0.1532 \\
			SAHeart      & 462  & 9   & 0.2332 & 0.2515 & 0.2027 & 0.1977 & \textbf{0.1959} & 0.2015 & 0.1928 & ... & ... & 0.2036 & 0.2090 & 0.2154 & 0.2157 \\
			Tic-Tac-Toe  & 958  & 9   & 0.2164 & 0.1777 & 0.2001 & 0.1950 & 0.2223 & 0.1747 & 0.2123 & ... & ... & 0.1116 & 0.0959 & 0.0881 & \textbf{0.0876} \\
			Heart        & 303  & 13  & 0.2499 & 0.1972 & 0.1482 & 0.1325 & 0.1461 & 0.2207 & \textbf{0.1268} & ... & ... & 0.1359 & 0.1392 & 0.1523 & 0.1400 \\
			House Vote   & 232  & 16  & 0.0690 & 0.0419 & 0.0405 & 0.0491 & 0.0512 & 0.1700 & 0.0617 & ... & ... & 0.0379 & \textbf{0.0364} & \textbf{0.0364} & \textbf{0.0364} \\
			Bands        & 365  & 19  & 0.2226 & 0.2461 & 0.2047 & 0.2117 & 0.2229 & 0.2473 & 0.2206 & ... & ... & 0.1820 & 0.1825 & 0.1810 & \textbf{0.1800} \\
			Parkinson    & 195  & 22  & 0.1515 & 0.1248 & 0.1193 & 0.1250 & 0.1396 & 0.1604 & 0.1218 & ... & ... & 0.1046 & 0.1022 & 0.0999 & \textbf{0.0989} \\
			Body         & 507  & 23  & 0.0328 & 0.0739 & 0.0582 & 0.0223 & \textbf{0.0174} & 0.2535 & 0.0400 & ... & ... & 0.0445 & 0.0411 & 0.0415 & 0.0401 \\
			Thyroid      & 9172 & 27  & 0.0328 & 0.0739 & 0.0582 & 0.0223 & 0.0174 & 0.2535 & 0.0400 & ... & ... & 0.0112 & 0.0411 & 0.0200 & \textbf{0.0110} \\
			WDBC         & 569  & 29  & 0.0384 & 0.0761 & 0.0462 & 0.0320 & 0.0309 & 0.0901 & 0.0402 & ... & ... & 0.0391 & 0.0390 & 0.0310 & \textbf{0.0300} \\
			WPBC         & 198  & 32  & 0.2017 & 0.2132 & 0.1800 & 0.1830 & 0.1810 & 0.1959 & 0.1862 & ... & ... & 0.1895 & 0.1899 & \textbf{0.1733} & 0.1745 \\
			Oil-Spill    & 937  & 49  & 0.0440 & 0.0387 & 0.0338 & 0.0349 & 0.0360 & 0.0417 & 0.0359 & ... & ... & 0.0329 & 0.0340 & 0.0322 & \textbf{0.0313} \\
			Spam base    & 4601 & 58  & 0.1682 & 0.0938 & 0.0914 & 0.0887 & 0.0676 & 0.2374 & 0.1375 & ... & ... & 0.0501 & 0.0477 & 0.0475 & \textbf{0.0471} \\
			Sonar        & 208  & 60  & 0.2296 & 0.2547 & 0.1758 & 0.1616 & 0.1905 & 0.2617 & 0.1919 & ... & ... & 0.1629 & 0.1638 & 0.1601 & \textbf{0.1582} \\
			Glaucoma     & 196  & 62  & 0.1648 & 0.1607 & 0.1261 & 0.1245 & 0.1389 & 0.2362 & 0.1422 & ... & ... & 0.1241 & \textbf{0.1153} & 0.1180 & 0.1194 \\
			Nki 70       & 144  & 76  & 0.1726 & \textbf{0.1507} & 0.1622 & 0.2006 & 0.2129 & 0.2393 & 0.2140 & ... & ... & 0.1879 & 0.1719 & 0.1816 & 0.1747 \\
			Musk         & 476  & 166 & 0.1625 & 0.2448 & 0.1844 & 0.1452 & 0.1681 & 0.2492 & 0.1876 & ... & ... & 0.1459 & 0.1413 & 0.1358 & \textbf{0.1337} \\
			\bottomrule
		\end{tabular}%
	\end{table}%
\end{landscape}

\begin{landscape}
	\renewcommand{\arraystretch}{2}
	\begin{table}[h]
		\caption{Sensitivity of $k$NN, tree, random forest, node harvest, SVM (with four different kernels), random projection with quadratic and linear discriminant analyses, OTE, OTE$_{oob}$ and OTE$_{sub}$. Results are based on 30\% training and 70\% testing parts of the data. Overall best performing method result is shown in bold.}
		\label{sen} %
		\setlength{\tabcolsep}{4pt}
		\begin{tabular}{lccccccccccccccc}
			\toprule
			Dataset & \multicolumn{1}{c}{$n$} & \multicolumn{1}{c}{$p$} & \multicolumn{1}{c}{kNN} & \multicolumn{1}{c}{Tree} & \multicolumn{1}{c}{NH} & \multicolumn{1}{c}{SVM} & \multicolumn{1}{c}{SVM} & \multicolumn{1}{c}{SVM} & \multicolumn{1}{c}{SVM} & \multicolumn{1}{c}{RP} & \multicolumn{1}{c}{RP} & \multicolumn{1}{c}{RF} & \multicolumn{1}{c}{OTE} & \multicolumn{1}{c}{OTE$_{oob}$} & \multicolumn{1}{c}{OTE$_{sub}$} \\
			&       &       &       &       &       & \multicolumn{1}{c}{(Radial)} & \multicolumn{1}{c}{(Linear)} & \multicolumn{1}{c}{(Bessel)} & \multicolumn{1}{c}{(Laplacian)} & \multicolumn{1}{c}{(LDA)} & \multicolumn{1}{c}{(QDA)} &       &       &       &  \\
			\midrule
			Mammographic & 830  & 5   & 0.8134 & 0.8160 & 0.8156 & 0.7953 & 0.8169 & 0.8345 & 0.8213 & 0.8414 & \textbf{0.8523} & 0.8186 & 0.8342 & 0.8331 & 0.8231 \\
			Dystrophy    & 209  & 5   & 0.6668 & 0.6407 & 0.6641 & 0.8039 & 0.8032 & 0.8028 & 0.8069 & \textbf{0.9569} & 0.9444 & 0.7316 & 0.7605 & 0.7513 & 0.7497 \\
			Monk3        & 122  & 6   & 0.8430 & 0.8981 & 0.8168 & 0.8021 & 0.7521 & 0.8146 & 0.7508 & 0.7697 & 0.8586 & 0.9109 & \textbf{0.9209} & 0.8755 & 0.8744 \\
			Appendicitis & 106  & 7   & 0.4325 & 0.4517 & 0.2537 & 0.5614 & 0.6269 & \textbf{0.6421} & 0.6206 & 0.5671 & 0.5368 & 0.5347 & 0.6041 & 0.4945 & 0.5249 \\
			SAHeart      & 462  & 9   & 0.3478 & 0.4438 & 0.3705 & 0.7248 & 0.6800 & 0.7075 & \textbf{0.7261} & 0.4703 & 0.4979 & 0.3932 & 0.4522 & 0.4448 & 0.4397 \\
			Tic-Tac-Toe  & 958  & 9   & 0.8749 & 0.8867 & 0.9202 & 0.6821 & 0.5832 & 0.7790 & 0.5195 & 0.9324 & 0.8837 & 0.9703 & 0.9847 & 0.9889 & \textbf{0.9899} \\
			Heart        & 303  & 13  & 0.5480 & 0.6700 & 0.7263 & 0.7928 & 0.7774 & 0.6357 & 0.7822 & 0.6906 & 0.7056 & 0.7596 & \textbf{0.7908} & 0.7633 & 0.7565 \\
			House Vote   & 232  & 16  & 0.9338 & 0.9535 & 0.9547 & 0.9273 & 0.9334 & 0.6278 & 0.9278 & 0.9447 & 0.9500 & 0.9506 & \textbf{0.9678} & 0.9447 & 0.9681 \\
			Bands        & 365  & 19  & 0.4722 & 0.5613 & 0.3675 & 0.4360 & 0.5508 & 0.4356 & 0.4766 & 0.4158 & 0.4928 & 0.5092 & 0.5923 & 0.5972 & \textbf{0.5985} \\
			Parkinson    & 195  & 22  & 0.9276 & 0.9148 & 0.9153 & 0.7969 & 0.7919 & 0.7965 & 0.7538 & 0.9260 & 0.9339 & 0.9328 & \textbf{0.9652} & 0.9372 & 0.9374 \\
			Body         & 507  & 23  & 0.9536 & 0.9201 & 0.9212 & 0.9653 & 0.9766 & 0.4272 & 0.9370 & 0.9889 & \textbf{0.9898} & 0.9369 & 0.9504 & 0.9355 & 0.9483 \\
			Thyroid      & 9172 & 27  & 0.9536 & 0.9201 & 0.9212 & 0.9653 & 0.9766 & 0.4272 & 0.9370 & 0.4954 & 0.4905 & 0.9289 & 0.9504 & 0.9786 & \textbf{0.9789} \\
			WDBC         & 569  & 29  & 0.9255 & 0.8702 & 0.9092 & 0.9452 & 0.9462 & 0.1921 & 0.9334 & 0.9844 & 0.9814 & 0.9191 & 0.9260 & 0.9864 & \textbf{0.9885} \\
			WPBC         & 198  & 32  & 0.1469 & 0.2880 & 0.1365 & 0.4031 & 0.4856 & 0.2937 & 0.3960 & 0.1370 & 0.2733 & 0.4563 & 0.4278 & 0.4993 & \textbf{0.5032} \\
			Oil-Spill    & 937  & 49  & 0.0167 & 0.3228 & 0.1027 & 0.5455 & 0.5907 & 0.1262 & 0.5372 & 0.8123 & 0.9491 & 0.1284 & 0.2241 & 0.9512 & \textbf{0.9537} \\
			Spam base    & 4601 & 58  & 0.6712 & 0.8266 & 0.8191 & 0.8760 & 0.9034 & 0.8750 & 0.6452 & 0.7983 & 0.8002 & 0.9057 & 0.9140 & 0.9235 & \textbf{0.9342} \\
			Sonar        & 208  & 60  & 0.5311 & 0.6461 & 0.6554 & 0.7285 & 0.6792 & 0.4473 & 0.5519 & 0.7145 & 0.7747 & 0.6908 & 0.7380 & \textbf{0.8100} & 0.8023 \\
			Glaucoma     & 196  & 62  & 0.7416 & 0.8333 & 0.8333 & 0.8067 & 0.8146 & 0.3765 & 0.7940 & 0.8278 & 0.8378 & 0.8247 & 0.8690 & 0.8727 & \textbf{0.8846} \\
			Nki 70       & 144  & 76  & 0.5387 & \textbf{0.6523} & 0.5788 & 0.5109 & 0.5139 & 0.0759 & 0.5000 & 0.5043 & 0.4350 & 0.4051 & 0.6383 & 0.6182 & 0.5939 \\
			Musk         & 476  & 166 & 0.9001 & 0.6237 & 0.6045 & 0.7515 & 0.7619 & 0.4515 & 0.4500 & 0.6763 & 0.7269 & 0.7470 & 0.7959 & \textbf{0.9203} & 0.9197 \\
			\bottomrule
		\end{tabular}%
	\end{table}%
\end{landscape}

\begin{landscape}
	\renewcommand{\arraystretch}{2}
	\begin{table}[h]
		\caption{Kappa values of $k$NN, tree, random forest, node harvest, SVM (with four different kernels), random projection with quadratic and linear discriminant analyses, OTE, OTE$_{oob}$ and OTE$_{sub}$. Results are based on 30\% training and 70\% testing parts of the data. Overall best performing method result is shown in bold.}
		\label{kappa} %
		\setlength{\tabcolsep}{4pt}
		\begin{tabular}{lccccccccccccccc}
			\toprule
			Dataset & \multicolumn{1}{c}{$n$} & \multicolumn{1}{c}{$p$} & \multicolumn{1}{c}{kNN} & \multicolumn{1}{c}{Tree} & \multicolumn{1}{c}{NH} & \multicolumn{1}{c}{SVM} & \multicolumn{1}{c}{SVM} & \multicolumn{1}{c}{SVM} & \multicolumn{1}{c}{SVM} & \multicolumn{1}{c}{RP} & \multicolumn{1}{c}{RP} & \multicolumn{1}{c}{RF} & \multicolumn{1}{c}{OTE} & \multicolumn{1}{c}{OTE$_{oob}$} & \multicolumn{1}{c}{OTE$_{sub}$} \\
			&       &       &       &       &       & \multicolumn{1}{c}{(Radial)} & \multicolumn{1}{c}{(Linear)} & \multicolumn{1}{c}{(Bessel)} & \multicolumn{1}{c}{(Laplacian)} & \multicolumn{1}{c}{(LDA)} & \multicolumn{1}{c}{(QDA)} &       &       &       &  \\
			\midrule
			Mammographic & 830  & 5   & 0.5619 & 0.6626 & \textbf{0.6653} & 0.6159 & 0.6390 & 0.6348  & 0.6273 & 0.6541 & 0.6259 & 0.6461 & 0.6350 & 0.5639 & 0.6082 \\
			Dystrophy    & 209  & 5   & 0.6534 & 0.5278 & 0.6432 & 0.7024 & 0.7220 & 0.7088  & 0.7273 & 0.7316 & \textbf{0.7385} & 0.6934 & 0.6803 & 0.6543 & 0.6722 \\
			Monk3        & 122  & 6   & 0.6139 & 0.7976 & 0.5729 & 0.6287 & 0.4858 & 0.6545  & 0.5121 & 0.5634 & 0.7882 & 0.8131 & \textbf{0.8145} & 0.7889 & 0.7789 \\
			Appendicitis & 106  & 7   & 0.4136 & 0.4875 & 0.1894 & 0.3934 & 0.3719 & 0.4414  & 0.4728 & \textbf{0.5399} & 0.4758 & 0.4875 & 0.4902 & 0.4223 & 0.4119 \\
			SAHeart      & 462  & 9   & 0.1585 & 0.1901 & 0.2331 & 0.3160 & 0.3147 & 0.2964  & 0.2996 & 0.2987 & 0.2899 & 0.2290 & 0.2455 & \textbf{0.3291} & 0.3188 \\
			Tic-Tac-Toe  & 958  & 9   & 0.1768 & 0.4254 & 0.2563 & 0.2741 & 0.1581 & 0.3935  & 0.0099 & 0.1756 & 0.4186 & 0.6778 & 0.7131 & 0.7352 & \textbf{0.7551} \\
			Heart        & 303  & 13  & 0.2263 & 0.4495 & 0.5796 & 0.6365 & 0.5998 & 0.2715  & \textbf{0.6406} & 0.5133 & 0.4953 & 0.6205 & 0.5921 & 0.5758 & 0.5912 \\
			House Vote   & 232  & 16  & 0.8078 & 0.8996 & 0.9003 & 0.8565 & 0.8633 & 0.4909  & 0.8156 & 0.8469 & 0.8509 & 0.9043 & 0.9028 & 0.9231 & \textbf{0.9341} \\
			Bands        & 365  & 19  & 0.2970 & 0.2928 & 0.2876 & 0.2694 & 0.2423 & -0.0263 & 0.0251 & 0.2791 & 0.3367 & 0.4149 & 0.4120 & 0.4158 & \textbf{0.4321} \\
			Parkinson    & 195  & 22  & 0.3584 & 0.6094 & 0.5058 & 0.4966 & 0.4719 & 0.3671  & 0.4517 & 0.5247 & 0.6399 & 0.5836 & 0.5701 & 0.6427 & \textbf{0.6479} \\
			Body         & 507  & 23  & 0.9121 & 0.8397 & 0.8498 & 0.9390 & 0.9595 & 0.0063  & 0.8870 & 0.9608 & \textbf{0.9650} & 0.8810 & 0.8850 & 0.8938 & 0.8968 \\
			Thyroid      & 9172 & 27  & 0.9121 & 0.8397 & 0.8498 & 0.9390 & 0.9595 & 0.0063  & 0.8870 & 0.5890 & 0.5893 & 0.8966 & 0.8850 & 0.9397 & \textbf{0.9687} \\
			WDBC         & 569  & 29  & 0.8965 & 0.8121 & 0.8697 & 0.9075 & 0.9133 & -0.5410 & 0.8808 & 0.9295 & 0.9225 & 0.8892 & 0.8909 & 0.9311 & \textbf{0.9352} \\
			WPBC         & 198  & 32  & 0.1023 & 0.1166 & 0.1089 & 0.1637 & 0.2021 & -0.0504 & 0.0181 & 0.1130 & 0.1524 & 0.0763 & 0.1295 & 0.1691 & \textbf{0.1699} \\
			Oil-Spill    & 937  & 49  & 0.0204 & 0.2946 & 0.1442 & 0.3419 & 0.3511 & -0.0202 & 0.1971 & 0.1464 & 0.3046 & 0.1867 & 0.2927 & 0.3604 & \textbf{0.3863} \\
			Spam base    & 4601 & 58  & 0.4918 & 0.7718 & 0.7940 & 0.7368 & 0.8307 & 0.1311  & 0.3918 & 0.7217 & 0.7232 & 0.8768 & 0.8815 & 0.8816 & \textbf{0.8817} \\
			Sonar        & 208  & 60  & 0.2875 & 0.3314 & 0.4719 & 0.5115 & 0.4578 & -0.0403 & 0.0295 & 0.4975 & 0.5875 & 0.5388 & 0.5043 & 0.\textbf{5965} & 0.5944 \\
			Glaucoma     & 196  & 62  & 0.5687 & 0.6182 & 0.6532 & 0.6596 & 0.6303 & -0.2283 & 0.4964 & 0.6921 & 0.6985 & 0.6704 & \textbf{0.7003} & 0.6856 & 0.6856 \\
			Nki 70       & 144  & 76  & 0.4693 & \textbf{0.5704} & 0.5146 & 0.2915 & 0.2559  & 0.0167 & 0.0004 & 0.2858 & 0.2441 & 0.3474 & 0.4922 & 0.4697 & 0.4697 \\
			Musk         & 476  & 166 & 0.5304 & 0.3504 & 0.4568 & 0.5503 & 0.5199 & 0.0038  & 0.0000 & 0.4241 & 0.5672 & 0.6129 & 0.6311 & \textbf{0.6472} & 0.6372 \\
			\bottomrule
		\end{tabular}%
	\end{table}%
\end{landscape}

 \twocolumn
 \subsection{Discussion} \label{disc}
 Results given in Tables \ref{results70}, \ref{results50} and \ref{results30} reveal that the OTE$_{oob}$ and OTE$_{sub}$ are almost always better than OTE.  The results in Table \ref{results70} also show that OTE with $V=10 \%$ is always giving better results than OTE with $V=15\%$ and so on, with the exception of Mammographic dataset only.  From Table \ref{results70}, that shows results based on 70\% and 30\% splitting of the data, it can be seen that node harvest and SVM gave better results than the others on 2 data sets each. RP ensemble gave better results than the rest on 4 datasets 2 each for LDA and  QDA base learner. OTE is better than the others on 2 datasets with $V=10\%$. OTE$_{oob}$ although better than OTE in most of the cases, outperformed the rest of the methods on 1 dataset. OTE$_{sub}$ gave better results than the others on 9 of the datasets. Tree and $k$NN methods could not outperformed the rest of the methods on any dataset.

 From Table \ref{results50}, that shows results based on 50\% and 50\% splitting of the data, it can be seen that node harvest and SVM gave better results than the others on 1 data sets each. RP ensemble gave better results than the rest on 6 datasets. Random forest is better than the others on 3 data sets. OTE is better than the others on 1 dataset. OTE$_{oob}$ is better than OTE in most of the cases, and outperformed the rest of the methods on 2 dataset. OTE$_{sub}$ gave better results than the others on 7 of the datasets. Tree and $k$NN methods could not outperformed the rest of the methods on any dataset.
 
 The results in Table \ref{results30}, based on 70\% and 30\% splitting of the data, show that node harvest and SVM gave better results than the others on 1 data sets each. RP ensemble gave better results than the rest on 4 datasets 2 each for LDA and  QDA base learner. Random forest is better than the others on 4 data sets. OTE could not outperformed the others on any of the data set. OTE$_{oob}$ is better than OTE in most of the cases, and outperformed the rest of the methods on 4 dataset. OTE$_{sub}$ gave better results than the others on 6 of the datasets. Tree and $k$NN methods could not outperformed the rest of the methods on any of the datasets.
 
Furthermore, the results of the methods in terms of Brier score, sensitivity and Kappa, given in Tables \ref{brier}, \ref{sen} and \ref{kappa}, respectively, indicate that the proposed ensembles outperformed the rest of the methods on majority of the datasets. Brier score values are not estimated for random projection ensemble (Table \ref{brier}) as the current implementation of the algorithm given in the R package \cite{RPEnsemble} does not support probability estimation.
 
\begin{figure}[htbp]
 	\centering
 	\includegraphics[width=10cm]{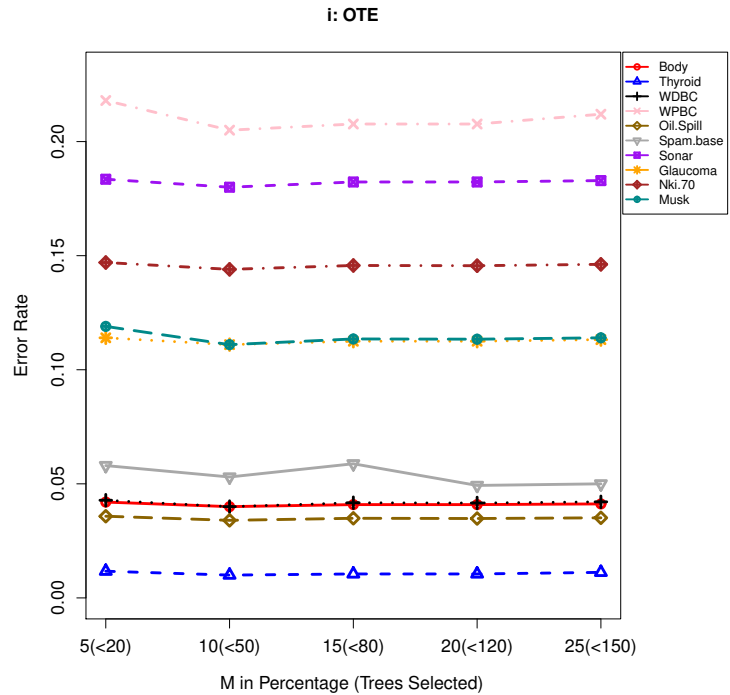} 
 	\caption{Effect of $M$ on the error rate of the data sets shown using \emph{OTE}. The value of $M$ in percentage is on the x-axis and error rate is on the y-axis. Number of trees selected are also shown in brackets on the x-axis, e.g. $10(<50)$ means that the method selected less than 50 trees for the datasets at $M=10\%$}
 	\label{a}
 \end{figure}

\begin{figure}[htbp]
 	\centering
 	 	\includegraphics[width=9.7cm]{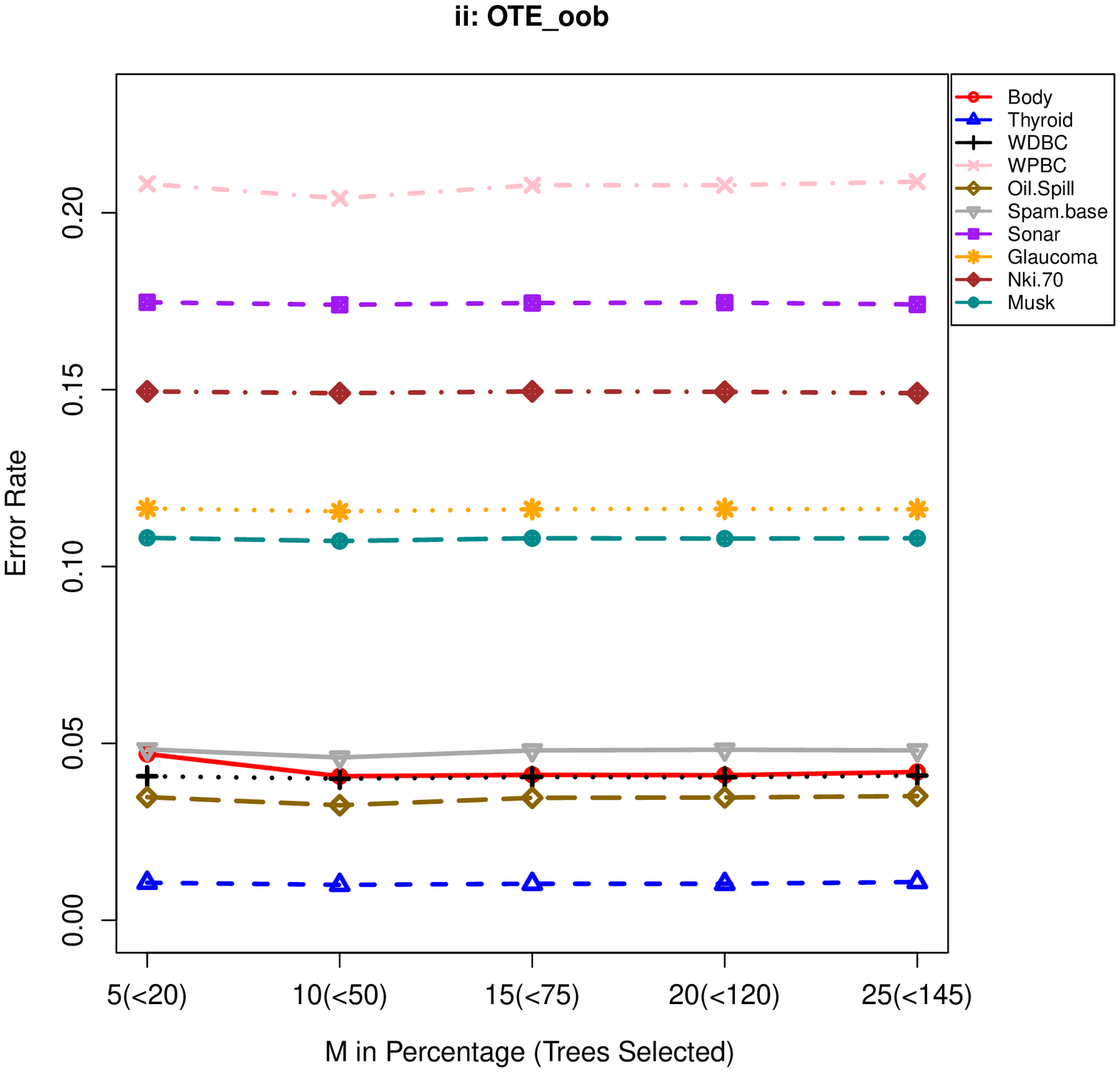} 
 	\caption{Effect of $M$ on the error rate of the data sets shown using \emph{OTE$_{oob}$}. The value of $M$ in percentage is on the x-axis and error rate is on the y-axis. Number of trees selected are also shown in brackets on the x-axis, e.g. $10(<50)$ means that the method selected less than 50 trees for the datasets at $M=10\%$}
 	\label{b}
 \end{figure}

\begin{figure}[htbp]
 	\centering
 	\includegraphics[width=10cm]{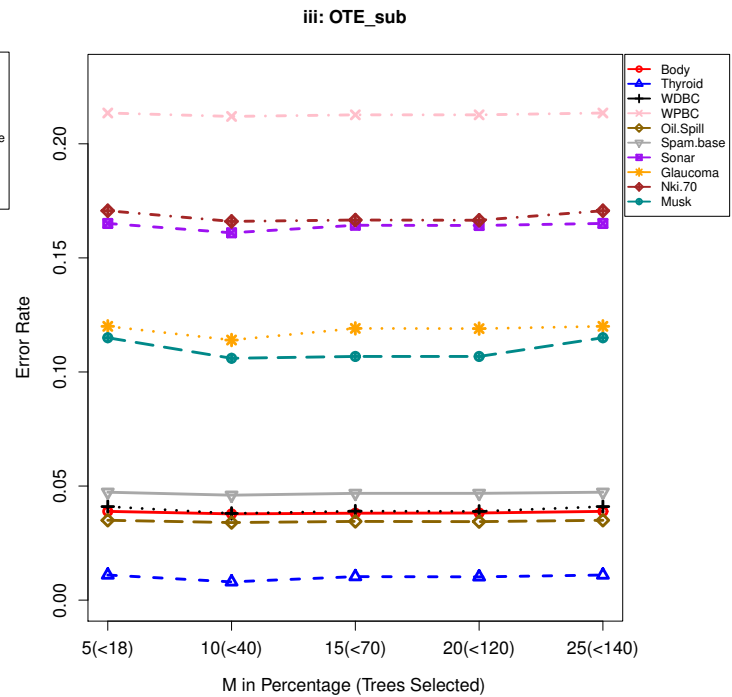} 
 	\caption{Effect of $M$ on the error rate of the data sets shown using \emph{OTE$_{oob}$}. The value of $M$ in percentage is on the x-axis and error rate is on the y-axis. Number of trees selected are also shown in brackets on the x-axis, e.g. $10(<40)$ means that the method selected less than 40 trees for the datasets at $M=10\%$}
 	\label{c}
 \end{figure}

 Moreover, the effect of choosing various number of trees on the three methods, i.e. OTE, OTE$_{oob}$ and OTE$_{sub}$ in terms of classification error rates are shown in Figure \ref{a}, \ref{b} and \ref{c}, respectively. In the given figures, the value of $M$ in percentage is shown on the x-axis and error rate on the y-axis. Number of trees selected are also shown in brackets on the x-axis, e.g. $10(<40)$ means that the method selected less than 40 trees for the datasets at $M=10\%$. 
 
 Getting  better/comparable results by using a forest of few accurate and diverse trees to those based on thousands of weak trees is encouraging in that this might reduce computational costs in terms of storage resources. From size assessment of the proposed ensemble methods, it is evident that they provide the best result with the number of trees less than 50. This is a clear reduction in the ensembles size and could have significant  practical implications. 
 \section{Conclusion} \label{conclusion}
 Two methods of selecting optimal trees, based on the individual strenght of a tree and trees collective performance, from an original ensemble of a large number of trees are proposed as an improvement to OTE. The selected trees are then combined together to vote for the class labels of the unseen data. Using as much as possible of the training data for growing trees in the two proposed method guarantees better results. This makes the trees individually strong and as the methods implement a diversity check on the trees while selecting them for the final ensemble, enough randomness is maintained in base learners meeting the basic principles of ensemble learning. The analyses given in the paper, both on simulated and benchmark datasets, revealed that the proposed methods outperform the other state-of-the-art methods. 
 
 R implementation of the proposed ensembles is given in Package ``\emph{OTE}''  \cite{ote}.
 
 The proposed ensemble in its current version takes more training time than the random forest algorithm. For example, with Thyriod data ($n= 9172, d=27$), the training times for random forest and the proposed methods were 4.56 and 6.41 seconds, respectively, on a 3 GHz Intel Core i7 computer with 8 GB memory running under mac OS X operating system. The methods proposed in the paper can model massive data with ultra high dimension using parallel computing as implemented in the R package \cite{parallel}, for example.  
 Using feature selection methods, \cite{mahmoud2014feature,prop,ahmed2020hybrid,wijaya2020stability,gao2020efficient,brahim2017ensemble,liu2020novel,li2017application}, might, in conjunction with the proposed ensembles, result in further improvements \cite{reddy2020analysis}. Using random projection approach as given in  \cite{RPEnsemble2017paper,RPEnsemble} with the tree selection methods proposed in this paper, may also give further improvements. The idea of classifier selection based on clustering (CSBS) \cite{parvin2015proposing}, for ensemble creation could also be used with the proposed ensembles for efficient results. For data sets with features measured on different scales, random forest with $P$-value adjusted split criteria can avoid biased feature selection within the tree algorithm \cite{lausen1994classification,lausen2004assessment,Potapov2012}
\bibliographystyle{unsrt}      
\bibliography{modified_ote}   
\EOD

\end{document}